%% file: ijcai26.tex
\documentclass{article}
\usepackage{ijcai26}

\usepackage{times}
\usepackage{soul}
\usepackage{url}
\usepackage[hidelinks]{hyperref}
\usepackage[utf8]{inputenc}
\usepackage[small]{caption}
\usepackage{graphicx}
\usepackage{amsmath}
\usepackage{amsthm}
\usepackage{booktabs}
\usepackage{algorithm}
\usepackage{algorithmic}
\usepackage[switch]{lineno}
\usepackage{amsfonts} 
\usepackage{multirow} 
\usepackage[table]{xcolor}
\usepackage{pifont}
\usepackage{tabularx}
\usepackage{booktabs}
\usepackage{array}
\usepackage{xcolor}
\usepackage{listings}

\colorlet{punct}{red!60!black}
\definecolor{background}{HTML}{F7F7F7} 
\definecolor{delim}{RGB}{20,105,176}
\colorlet{numb}{magenta!60!black}

\lstdefinelanguage{json}{
    basicstyle=\ttfamily\scriptsize, 
    numbers=left,
    numberstyle=\tiny,
    stepnumber=1,
    numbersep=8pt,
    showstringspaces=false,
    breaklines=true,      
    breakatwhitespace=false,  
    frame=single,         
    backgroundcolor=\color{background}, 
    literate=
     *{:}{{{\color{punct}{:}}}}{1}
      {,}{{{\color{punct}{,}}}}{1}
      {\{}{{{\color{delim}{\{}}}}{1}
      {\}}{{{\color{delim}{\}}}}}{1}
      {[}{{{\color{delim}{[}}}}{1}
      {]}{{{\color{delim}{]}}}}{1},
}
\newcolumntype{Y}{>{\raggedright\arraybackslash}X}

\title{Muse: A Multi-agent Framework for Unconstrained Story Envisioning via Closed-Loop Cognitive Orchestration}

\author{
Wenzhang Sun$^{*1}$
\and
Zhenyu Wang$^{*1}$\and
Zhangchi Hu$^{2}$\and
Chunfeng Wang$^1$\and
Hao Li$^1$\and
Wei Chen$^1$\\
\affiliations
$^1$Li Auto.\
$^2$University of Science and Technology of China\\
\url{https://sunwenzhang1996.github.io/MUSE/}
}

\begin{document}
\maketitle
\let\thefootnote\relax\footnotetext{* Equal contribution.}
\setcounter{footnote}{0}
\input{sections/abs}    
\input{sections/intro}

\input{sections/related_work}

\input{sections/method}
\input{sections/benchmark}

\input{sections/experiments}

\input{sections/conclusion}
\bibliographystyle{named}
\bibliography{ijcai26}
\input{sections/supp}
\end{document}

%% file: sections/abs.tex
\begin{abstract}
Generating long-form audio-visual stories from a short user prompt remains challenging due to an intent–execution gap, where high-level narrative intent must be preserved across coherent, shot-level multimodal generation over long horizons. Existing approaches typically rely on feed-forward pipelines or prompt-only refinement, which often leads to semantic drift and identity inconsistency as sequences grow longer.
We address this challenge by formulating storytelling as a closed-loop constraint enforcement problem and propose MUSE, a multi-agent framework that coordinates generation through an iterative plan–execute–verify–revise loop. MUSE translates narrative intent into explicit, machine-executable controls over identity, spatial composition, and temporal continuity, and applies targeted multimodal feedback to correct violations during generation.
To evaluate open-ended storytelling without ground-truth references, we introduce MUSEBench, a reference-free evaluation protocol validated by human judgments. Experiments demonstrate that MUSE substantially improves long-horizon narrative coherence, cross-modal identity consistency, and cinematic quality compared with representative baselines.
\end{abstract}

%% file: sections/intro.tex
\section{Introduction}
\begin{figure}[t]
    \centering
    \includegraphics[width=0.99\linewidth]{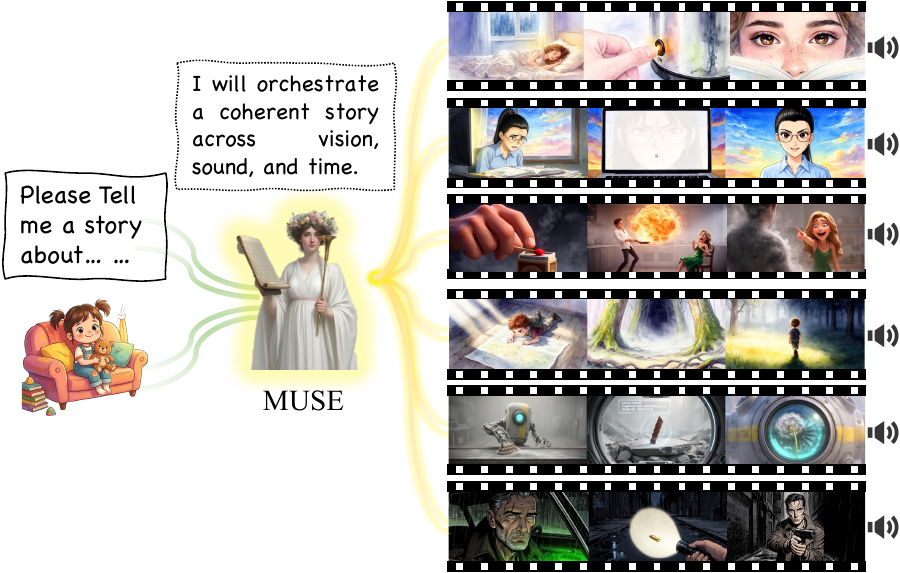}
    \caption{With only simple text inputs, MUSE can generate storytelling videos of diverse styles with high continuity.}
    \label{fig:teaser}
    \vspace{-4mm}
\end{figure}

Producing a coherent long-form audio-visual story from a short input remains a fundamental yet unsolved challenge in multimodal generation. Unlike short video synthesis, long-horizon storytelling requires a system to preserve global narrative intent—such as character identity, spatial relationships, cinematic composition, and causal event progression—across a sequence of shots that may span dozens of generation steps. As the generation horizon grows, even minor local deviations can accumulate, resulting in semantic drift and fragmented narratives that break the viewer’s immersion.

Most existing text-to-video and multimodal storytelling systems address this challenge in a largely feed-forward manner. A high-level prompt or script is first expanded into a sequence of textual descriptions, which are then independently rendered into visual and audio outputs. Recent agentic or prompt-based approaches partially mitigate errors, but still lack explicit mechanisms for enforcing global narrative constraints. Consequently, long-form generation commonly exhibits recurring failure modes, including cross-shot identity drift, spatial and cinematic inconsistency, and temporal discontinuity between adjacent scenes.
We argue that these failures are not merely due to imperfect generation models, but stem from a deeper intent–execution gap. Narrative intent is specified at an abstract, symbolic level, while execution is delegated to stochastic multimodal generators that operate locally and myopically. Without persistent, machine-interpretable constraints, high-level intent cannot be reliably enforced over long horizons. This gap becomes especially pronounced in audio-visual storytelling, where coherence must be maintained not only within each modality, but also across modalities and time.

To bridge this gap, we propose to view long-form audio-visual storytelling not as a single-pass generation problem, but as a closed-loop constraint enforcement process. From this perspective, narrative intent should be explicitly planned, continuously verified against generated outputs, and corrected whenever violations are detected. Such a formulation shifts the focus from unconstrained generation toward controllable and auditable storytelling, where coherence emerges from iterative enforcement rather than chance alignment.

To this end, we introduce MUSE, a multi-agent framework for long-form audio-visual storytelling. MUSE decouples high-level planning from low-level execution and coordinates generation through an iterative plan–execute–verify–revise loop. Instead of relying on natural-language prompt retries, MUSE represents narrative intent as explicit, machine-executable controls over key aspects of storytelling, including character identity, spatial composition, and temporal continuity. Generated visual and audio outputs are then analyzed to detect structured multimodal violations, enabling targeted and bounded revisions rather than unconstrained resampling. This design allows MUSE to maintain global narrative consistency while preserving the diversity and expressiveness of underlying generative models. Unlike prior agentic or iterative generation frameworks that treat verification as heuristic self-refinement, MUSE formulates long-form storytelling as an explicit constraint enforcement problem, where narrative intent is represented as machine-executable constraints and violations are detected and corrected through structured, typed multimodal signals.

Evaluating open-ended storytelling presents an additional challenge, as long-form narratives typically lack ground-truth references. To address this, we further introduce MUSEBench, a reference-free evaluation protocol that assesses narrative coherence and cross-modal identity consistency using large multimodal model–based scoring, and validate its reliability through human judgment studies. MUSEBench enables systematic comparison of storytelling systems without restricting generation to predefined scripts or templates. 

We evaluate MUSE across diverse storytelling scenarios and compare it with representative feed-forward and agentic baselines. Experimental results show that MUSE substantially improves long-horizon narrative coherence, cross-modal identity consistency, and cinematic quality, demonstrating the effectiveness of closed-loop constraint enforcement for long-form audio-visual storytelling. In summary, our contributions are threefold:

(1) We reformulate long-form audio-visual storytelling as a closed-loop constraint enforcement problem, explicitly modeling narrative intent as machine-executable constraints to bridge the intent–execution gap between high-level prompts and reliable shot-level generation over long horizons.

(2) We propose MUSE, a multi-agent framework that enforces global narrative consistency through a structured plan–execute–verify–revise loop, using explicit control representations and typed multimodal feedback to enable targeted, bounded corrections across vision, audio, and time.

(3) We introduce MUSEBench, a reference-free, multi-dimensional evaluation protocol for open-ended audio-visual storytelling, and validate its reliability through human–metric alignment studies, enabling holistic assessment beyond reference-dependent benchmarks.

%% file: sections/related_work.tex
\section{Related Work}
\begin{figure*}[t]
    \centering
    \includegraphics[width=0.99\linewidth]{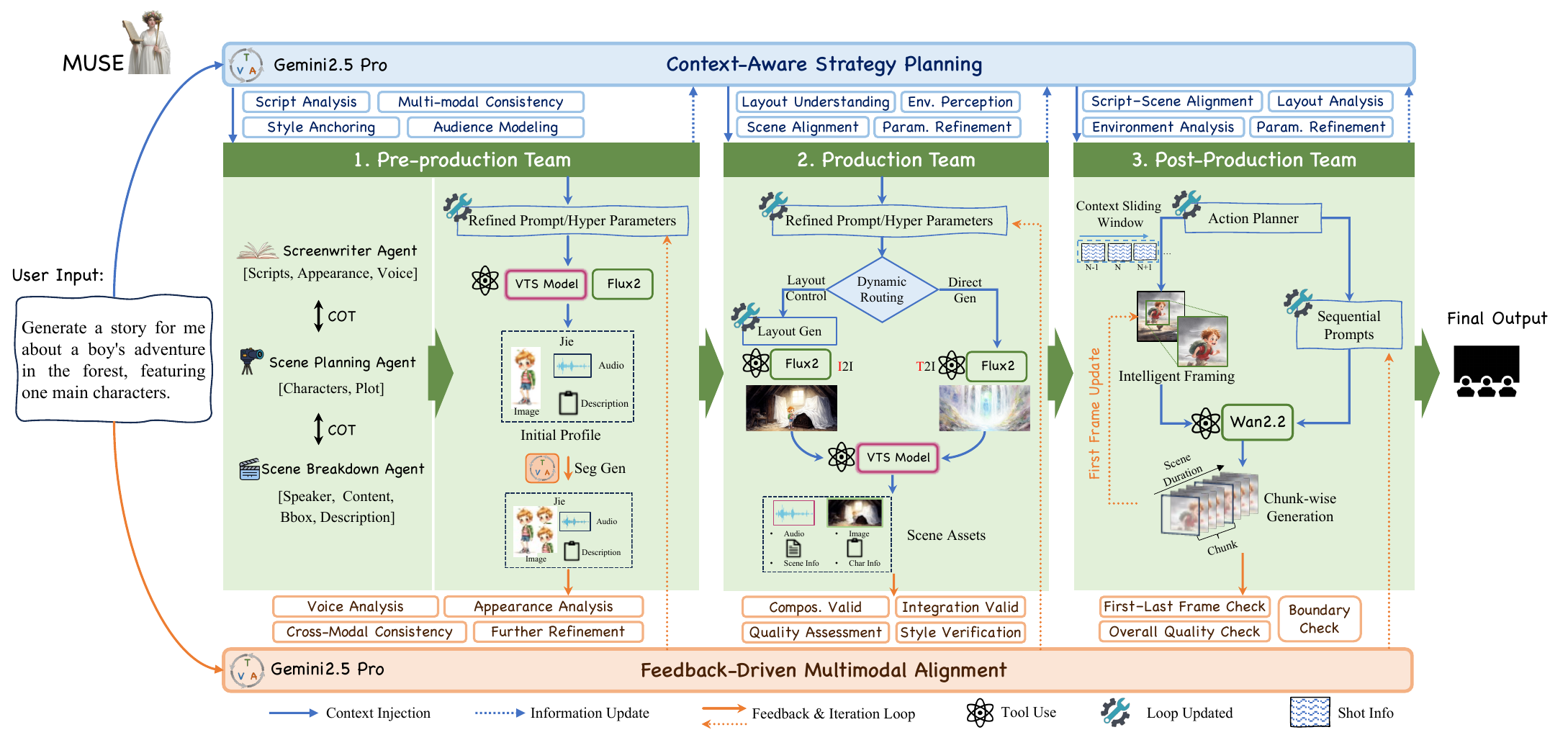}
    \vspace{-2mm}
    \caption{Overview of MUSE. Long-form audio-visual storytelling is realized through a closed-loop orchestration that coordinates specialist agents across identity (pre-production), space (production), and time (post-production).}
    \label{fig:overview}
    \vspace{-3mm}
\end{figure*}
Our work lies at the intersection of long-form video generation, agentic planning, and semantics-driven audio synthesis.

\noindent\textbf{Narrative Consistency in Long-form Video Generation.}
Recent diffusion-based text-to-video models have demonstrated impressive visual quality for short clips \cite{liu2024sora,ho2022imagen,ren2025videoworld,singer2022make,2505.16819,wang2025mavis,listory2screen,yin2025structure}.
However, these models largely operate in a feed-forward manner, making it difficult to preserve narrative coherence over long temporal horizons \cite{liu2025hi}.
As a result, identity drift and semantic inconsistency frequently emerge in extended sequences \cite{villegas2022phenaki,wu2022nuwa,elmoghany2025survey,waseem2025video,zhang2025semantic,liu2025moee}.
Prior efforts address this issue by introducing architectural or inference-time mechanisms such as sliding-window attention or reference-based conditioning \cite{zhou2024storydiffusion,ren2024consisti2v,guo2023animatediff,yin_grpose}.
While effective for short-term visual consistency, these approaches lack explicit mechanisms for enforcing high-level narrative logic and causal dependencies derived from scripts.
In contrast, MUSE treats long-form video generation as a structured planning problem and explicitly manages narrative constraints through closed-loop orchestration.

\noindent\textbf{Language Agents and Planning.}
Large Language Models have enabled agentic systems capable of reasoning, planning, and iterative self-improvement \cite{junlin2025large,achiam2023gpt,bubeck2023sparks}.
Multi-agent frameworks have been successfully applied to collaborative problem solving in domains such as software development \cite{hong2023metagpt,qian2024chatdev} and creative text generation \cite{wu2025automated,hu2024storyagent,kangcharacter}.
However, most existing creative agents remain limited to textual outputs or static representations, and their execution pipelines are typically open-loop, lacking mechanisms to verify whether generated multimodal content faithfully aligns with the intended semantics.
MUSE extends agentic planning into the multimodal domain by coupling structured planning with visual and audio verification, enabling targeted correction when constraint violations occur during generation.

\noindent\textbf{Semantics-Driven Zero-Shot Audio Synthesis.}
Conventional TTS and voice cloning systems rely on reference audio to establish vocal identity \cite{wang2023neural,ju2024naturalspeech,casanova2022yourtts}.
Although effective for imitation, this paradigm is misaligned with creative storytelling scenarios, where users often describe voices using abstract semantic attributes rather than concrete audio samples.
Recent efforts explore text-conditioned or generative audio models \cite{liu2023audioldm,guo2023prompttts,liu2025audiobook,mannonov2025bridging}, but consistent character-level voice control without references remains challenging.
Our Vocal Trait Synthesis (VTS) module addresses this gap by deriving stable vocal representations directly from semantic descriptions, enabling reference-free and identity-consistent speech generation for long-form narratives.
\input{tables/methodoverview}

%% file: tables/methodoverview.tex
\begin{table}[t]
\centering
\caption{\textbf{Capability checklist (from reported features in prior papers / released code).}
Compared with representative storytelling agents, MUSE additionally supports customized character voices for consistent audio-visual identity.}
\vspace{-2mm}
\label{tab:capabilities}
\resizebox{0.95\linewidth}{!}{
\begin{tabular}{l|c c c c}
\toprule
\textbf{Method} & \textbf{Visual Consist.} & \textbf{Long Script} & \textbf{Audio Narration} & \textbf{Customized Voice} \\
\midrule
V-GOT          & \checkmark & \checkmark & \ding{55}  & \ding{55} \\
MovieAgent     & \checkmark & \checkmark & \ding{55}  & \ding{55} \\
Anim-Director  & \checkmark & \checkmark & \ding{55}  & \ding{55} \\
MM-StoryAgent  & \checkmark & \checkmark & \checkmark & \ding{55} \\
\textbf{MUSE (Ours)} & \textbf{\checkmark} & \textbf{\checkmark} & \textbf{\checkmark} & \textbf{\checkmark} \\
\bottomrule
\end{tabular}}
\vspace{-4mm}
\end{table}

%% file: sections/method.tex
\section{Method}
\label{sec:method}

\subsection{Problem Definition}
\label{sec:task_def}

We study long-form audio-visual storytelling from a short user prompt. Given an unstructured prompt $\mathcal{U}$, the goal is to generate a sequence of shots $\mathcal{V}=\{v_1,\dots,v_N\}$ that realizes the intended narrative while maintaining global consistency over long horizons. We formulate storytelling as satisfying a set of global constraints $\mathcal{C}$, including narrative integrity, cross-modal character identity, spatial and cinematic coherence, and temporal continuity.
Direct feed-forward mapping from $\mathcal{U}$ to $\mathcal{V}$ is prone to error accumulation. To enable controllable generation, MUSE expands $\mathcal{U}$ into a structured script $\mathcal{S}=\{s_1,\dots,s_N\}$, where each segment $s_i$ specifies visual intent $\mathbf{I}_i$ (characters, scene, camera) and audio intent $\mathbf{A}_i$ (narration or dialogue). Unlike prior agentic storytelling systems, narrative intent is explicitly represented as enforceable constraints that persist across generation steps, rather than being implicitly encoded in prompts or planner states.

\subsection{Closed-Loop Omni-Modal Orchestration}
\label{sec:orchestration}

MUSE is a modular multi-agent system coordinated by an omni-modal controller $\mathcal{M}$. Rather than treating generation as a single-pass process, MUSE formulates long-form storytelling as an iterative plan–execute–verify–revise loop.

\noindent\textbf{Global memory and executable controls.}
MUSE maintains a shared state memory $\mathcal{H}$ that stores persistent information across generation steps, including character identities, shot-level constraints, accepted layouts, synthesized audio, and terminal states. For each script segment $s_i$ and agent $k$, the controller produces a structured control bundle $\Theta_{i,k}$ (e.g., identity anchors, layouts, routing decisions, and temporal boundaries), explicitly separating narrative intent from machine-executable controls.

\noindent\textbf{Unified closed-loop execution.}
For each segment $s_i$, MUSE iteratively refines generation through:
\begin{align}
\Theta_{i,k}^{(t)} &= \Phi_k(s_i,\mathcal{H}^{(t)}), \\
x_{i,k}^{(t)} &= \texttt{Agent}_k(\Theta_{i,k}^{(t)}), \\
\mathbf{e}_{i,k}^{(t)} &= \Psi_k(x_{i,k}^{(t)}, s_i, \mathcal{H}^{(t)}), \\
\mathcal{H}^{(t+1)},\Theta_{i,k}^{(t+1)} &= \Omega_k(\mathcal{H}^{(t)},\Theta_{i,k}^{(t)},\mathbf{e}_{i,k}^{(t)}),
\end{align}
where $\Phi$ produces executable controls, $\Psi$ performs structured multimodal verification, and $\Omega$ applies targeted revisions.
Accepted outputs are committed to $\mathcal{H}$ and reused by subsequent segments, preventing silent accumulation of inconsistencies.

\noindent\textbf{Difference from generic self-refinement.}
Unlike prompt-only retries, MUSE operates on structured controls rather than natural-language prompts, and feedback is expressed as typed violation signals (e.g., identity mismatch, layout violation, temporal leakage) with localized corrective actions.
This enables bounded, targeted revisions instead of unconstrained resampling.

\subsection{Script Decomposition and Identity Anchoring}
\label{sec:phase1}
The pre-production phase establishes \emph{global narrative states} that must remain invariant throughout the story, including shot structure and character identity.
This phase functions as a global planner that converts an abstract script into executable identity constraints before any visual or audio rendering.

\begin{figure}[t]
    \centering
    \includegraphics[width=0.99\linewidth]{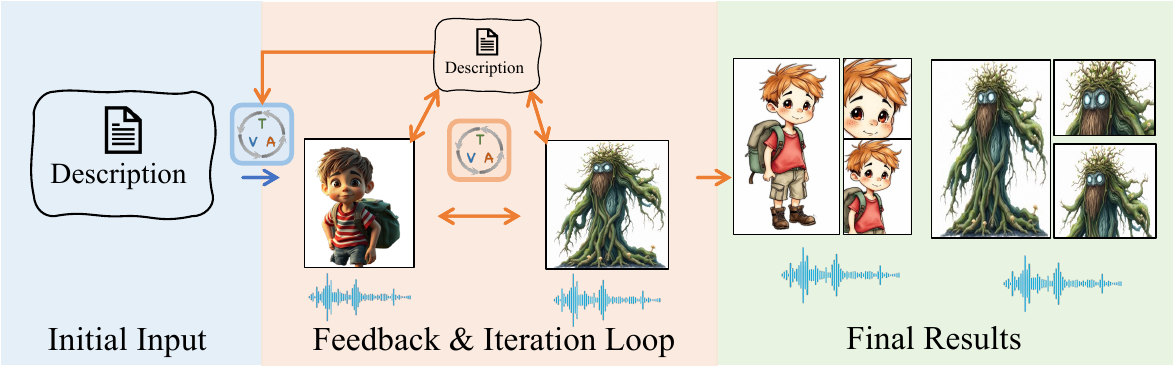}
    \vspace{-2mm}
    \caption{MUSE can generate diverse character assets from text descriptions while ensuring cross audio-visual consistency and inter-character identity distinction.}
    \label{fig:charassetgen}
    \vspace{-4mm}
\end{figure}

\paragraph{Planning: script structuring and identity state construction ($\Phi_{\text{pre}}$).}
Given the intermediate script $\mathcal{S}$, the planner first decomposes it into an ordered sequence of shots and extracts the set of participating characters.
For each character $c$, MUSE constructs a persistent multimodal identity state:
\vspace{-2mm}
\begin{equation}
\mathbf{z}_c = \{\mathbf{z}^{(c)}_{\text{vis}}, \mathbf{z}^{(c)}_{\text{voc}}\},
\end{equation}
which is stored in shared memory $\mathcal{H}$ and reused across all subsequent stages (Figure \ref{fig:charassetgen}).

The visual anchor $\mathbf{z}^{(c)}_{\text{vis}}$ is derived by synthesizing reference character assets under explicit appearance constraints (e.g., age, attire, and style descriptors), ensuring that downstream generators are conditioned on a stable identity representation rather than free-form prompts.
In parallel, MUSE introduces Vocal Trait Synthesis (VTS) to construct a vocal anchor $\mathbf{z}^{(c)}_{\text{voc}}$ directly from semantic descriptors such as age, gender, timbre, and speaking style.
This design locks acoustic identity \emph{prior} to generation, eliminating the need for reference audio and preventing voice drift across scenes.
\begin{figure}[t]
    \centering
    \includegraphics[width=0.99\linewidth]{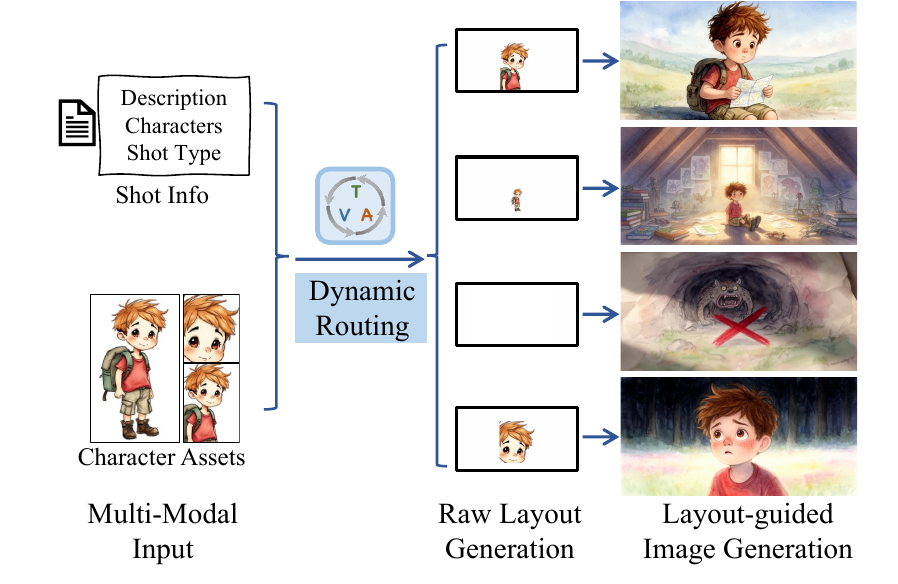}
    \vspace{-2mm}
    \caption{Dynamic routing enables the generation of diverse camera-movement shots while preserving identity stability.}
    \label{fig:dynamicrouting}
    \vspace{-4mm}
\end{figure}

\noindent\textbf{Verification and revision: identity consistency enforcement ($\Psi_{\text{pre}}, \Omega_{\text{pre}}$).}
The feedback stream verifies identity correctness at two levels.
First, it checks \emph{instruction alignment}, ensuring that each visual and vocal asset is semantically consistent with corresponding descriptors.
Second, it evaluates inter-character consistency, verifying alignment across distinct character assets. Upon detecting violations, the revision module performs targeted regeneration of conflicting modalities while preserving the remaining identity states—preventing early identity errors from propagating to subsequent shots.

\subsection{Layout-Aware Multimodal Asset Synthesis}
\label{sec:phase2}

With global identities fixed, the production phase synthesizes shot-level visual and audio assets while enforcing spatial and cinematic constraints.
The key challenge is to translate script-level composition into reliable pixel-level structure.

\paragraph{Planning: routing and spatial control ($\Phi_{\text{prod}}$).}
For each shot $s_i$, MUSE selects an execution route based on scene asset (Figure \ref{fig:dynamicrouting}).
Rather than committing to a single generator, the planner dynamically chooses between direct generation and layout-guided synthesis, enabling explicit control when multiple entities or camera constraints are present. When spatial control is required, $\Phi_{\text{prod}}$ synthesizes a coarse layout
$L^{(i)}_{\text{bbox}}$ that specifies the approximate position and scale of characters.
This layout is injected as a hard structural prior into the visual generation backbone, ensuring that entity presence and composition are respected (Figure \ref{fig:layoutgen}).
To avoid degenerate cases where entities collapse to low-resolution regions, a geometric guardrail enforces minimum spatial extents and resolves severe overlaps. In parallel, the audio agent generates narration or dialogue conditioned on the frozen vocal anchors $\mathbf{z}_{\text{voc}}$, ensuring that speech remains consistent in timbre and speaking style across all shots.

\begin{figure}[t]
    \centering
    \includegraphics[width=0.99\linewidth]{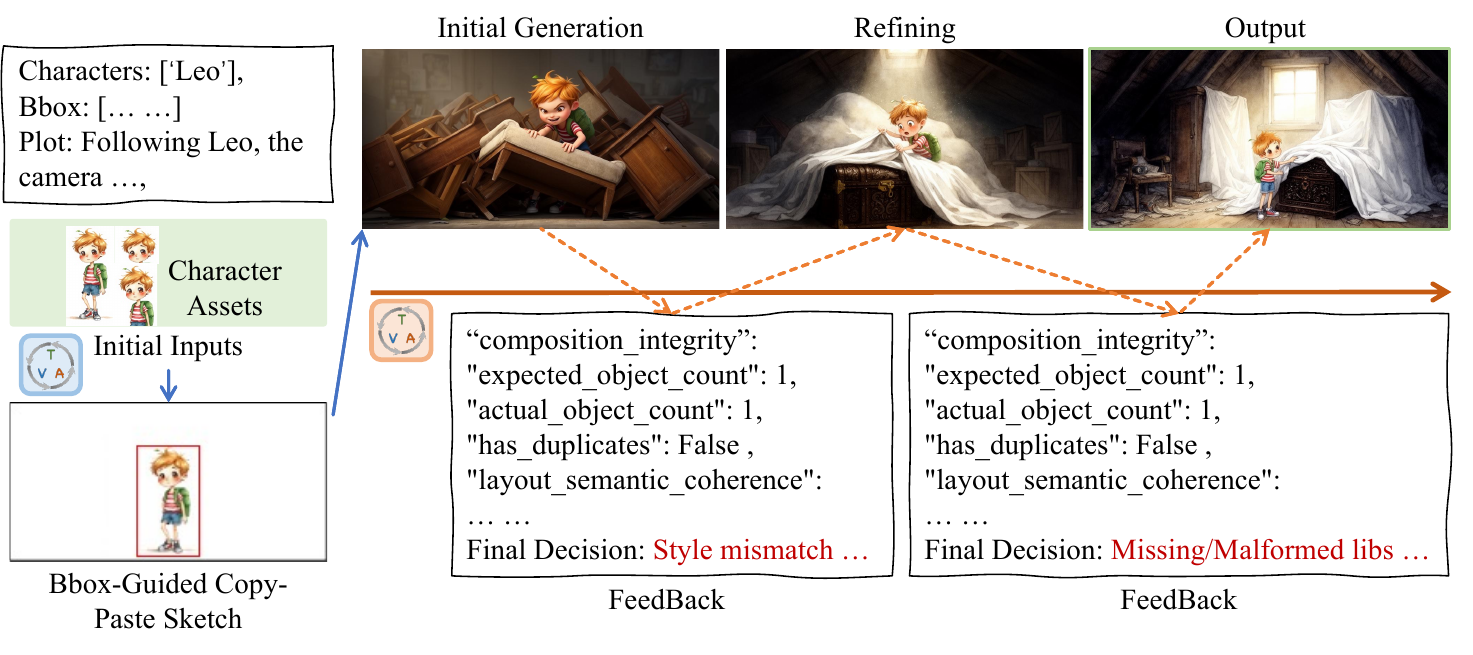}
    \vspace{-2mm}
    \caption{The feedback module evaluates generated scene assets from multiple perspectives and provides revision suggestions.}
    \label{fig:scenefeedback}
    \vspace{-4mm}
\end{figure}

\noindent\textbf{Verification and revision: spatial and compositional integrity ($\Psi_{\text{prod}}, \Omega_{\text{prod}}$).}
The feedback stream evaluates whether generated assets satisfy spatial constraints and integration quality (Figure \ref{fig:scenefeedback}).
Specifically, it checks (i) entity presence and (ii) visual coherence, including lighting consistency and the absence of compositional artifacts introduced by asset integration.
Upon detecting violations, the revision operator applies localized corrections, such as adjusting layouts, modifying generation routes, or refining guidance configurations.
Crucially, revisions are constrained to the identified failure regions, avoiding unconstrained re-generation of the entire shot.
\subsection{Temporal Synthesis}
\label{sec:phase3}

The post-production phase assembles static multimodal assets into temporally coherent video shots while enforcing narrative boundaries between segments.
\begin{figure*}[t]
    \centering
    \includegraphics[width=0.99\linewidth]{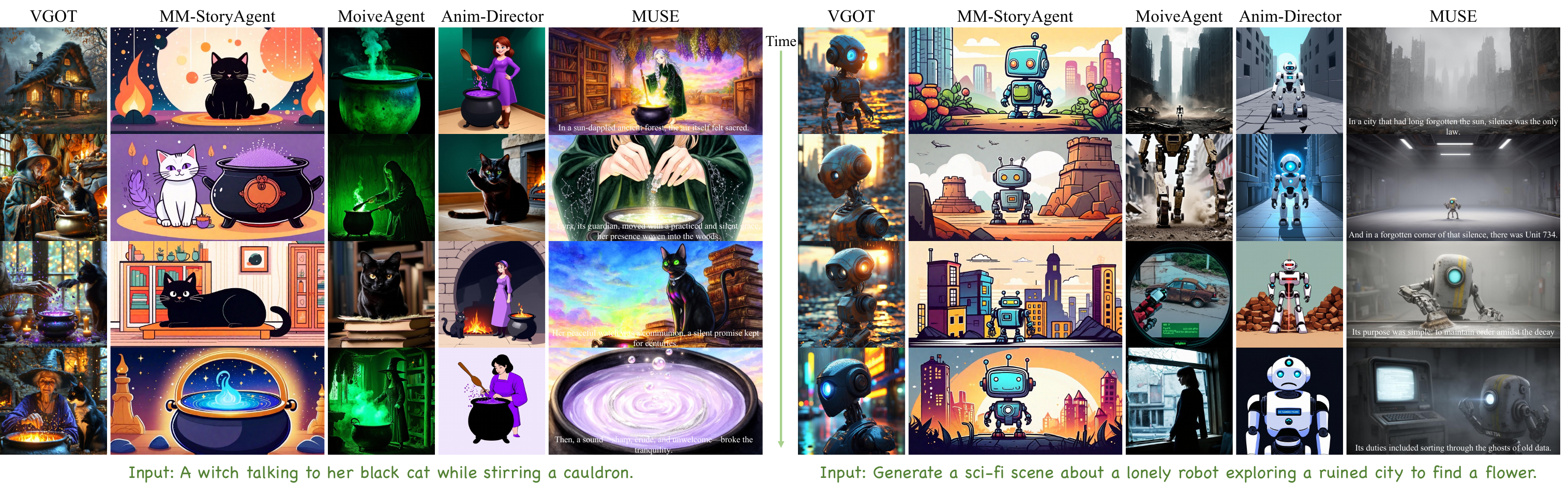}
    \vspace{-2mm}
    \caption{We show the first four consecutive shots of two narratives: Story A (Fantasy/Witch) and Story B (Sci-Fi/Robot). MUSE demonstrates strong style versatility across genres, robust identity persistence for the non-human protagonist in Story B (where baselines exhibit structural hallucination), and diverse cinematic framing via dynamic camera angles, outperforming baselines with static compositions.}
    \label{fig:mainfigure}
    \vspace{-4mm}
\end{figure*}
\noindent\textbf{Planning: temporal state propagation ($\Phi_{\text{post}}$).}
Independent generation of video shots often leads to state resets, causing motion discontinuities and broken action logic.
To address this, MUSE models temporal generation as a state-conditioned process.
Each shot $v_i$ is generated by conditioning on both the current script segment $s_i$ and a compact representation of the terminal state of the previous shot:
\begin{equation}
v_i = \texttt{VideoGen}\big(s_i \mid \texttt{Tail}(v_{i-1}), \mathbf{z}_{\text{vis}}\big),
\end{equation}
where $\texttt{Tail}(v_{i-1})$ encodes final-frame visual and motion cues.
An action planner further specifies shot-level temporal controls, including camera motion, actor motion, and duration, which are injected as explicit constraints.

\noindent\textbf{Verification: continuity and boundary compliance ($\Psi_{\text{post}}, \Omega_{\text{post}}$).}
After generation, the feedback stream verifies temporal continuity and boundary correctness.
It checks whether motion transitions between consecutive shots are plausible and whether the terminal frames of $v_i$ satisfy the end-state implied by $s_i$.
If a boundary violation is detected, the revision operator either truncates the tail frames or regenerates the shot under stricter temporal constraints.
This ensures that narrative progression remains well-structured and prevents semantic leakage across scene boundaries.

%% file: sections/benchmark.tex
\section{MUSEBench}
\label{sec:benchmark}
\begin{figure}[t]
    \centering
    \includegraphics[width=0.99\linewidth]{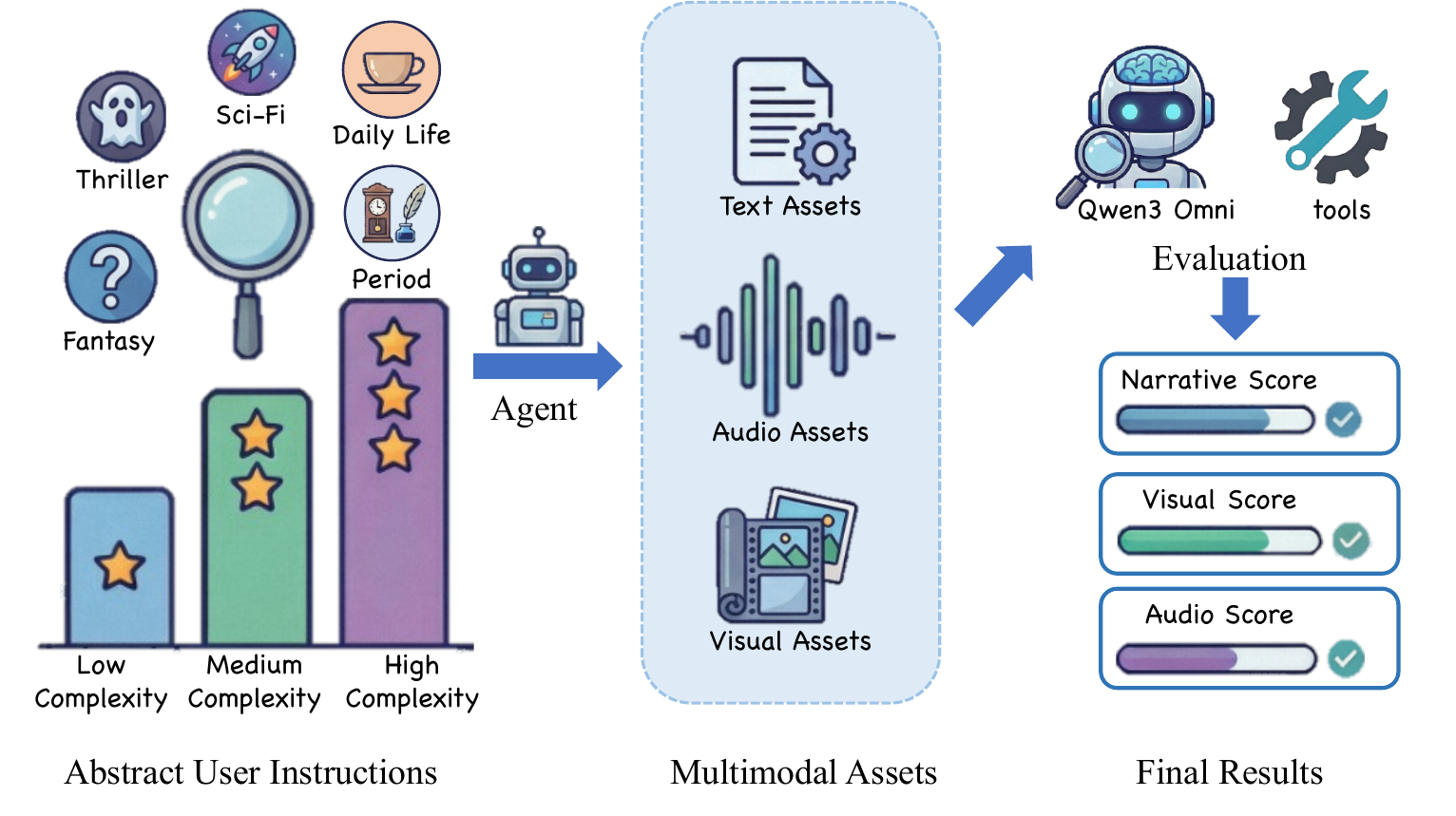}
    \vspace{-2mm}
    \caption{Overall pipeline of MUSEBench.
MUSEBench adopts an open-ended evaluation paradigm to assess the overall capabilities of storytelling systems from multiple complementary perspectives.}
    \label{fig:layoutgen}
    \vspace{-4mm}
\end{figure}
\paragraph{Motivation.} 
Recent benchmarks for story video generation, such as ViStoryBench \cite{zhuang2025vistorybench} and VinaBench \cite{2503.20871}, have made significant strides, particularly in quantifying visual consistency. However, these protocols typically rely on pre-defined intermediate assets (e.g., ground-truth character images or scripts) as evaluation anchors. 
For an end-to-end storytelling agent, such reliance is limiting; it restricts evaluation to isolated sub-modules rather than assessing the agent's holistic orchestration capabilities—specifically its proficiency in autonomous script decomposition, audio-visual synthesis, and cross-modal alignment. 
To bridge this gap, we introduce \textbf{MUSEBench}, an open-ended benchmarking framework. Unlike traditional metrics, MUSEBench takes only simple, abstract user prompts as input and rigorously evaluates both the \textit{generated intermediate reasoning} (e.g., script logic, narrative state) and the \textit{final multimedia output} (audio fidelity, visual aesthetics) across multiple dimensions.

MUSEBench encompasses 30 curated narrative prompts designed to challenge agentic generation limits (Figure \ref{sec:benchmark}). It spans five genres (\textit{Thriller, Daily Life, Period Piece, Science Fiction, Fantasy}) to test stylistic versatility, and covers a complexity spectrum ranging from single-character scenes to intricate multi-agent interactions with dynamic state changes. Several metrics in MUSEBench are evaluated using Large Multimodal Models (LMMs). We validate the correlation between these metrics and human preferences to demonstrate the effectiveness of MUSEBench.
While automatic evaluation of open-ended storytelling is inherently challenging, our goal is not to replace human judgment, but to provide a scalable proxy that correlates with it. Detailed evaluation metrics are documented in the Supp. 4.

%% file: sections/experiments.tex
\section{Experiments}
\label{sec:experiments}

\subsection{Experimental Setup}

\noindent\textbf{Implementation Details.}
MUSE is implemented as a modular multi-agent system following the closed-loop orchestration in Sec.~\ref{sec:orchestration}.
The cognitive backbone employs Gemini2.5 Pro\cite{comanici2025gemini} for reasoning and critique.
Visual synthesis uses Flux.2-Dev \cite{flux-2-2025} for image/asset generation and Wan2.2-I2V-A14B \cite{2503.20314} for video chunk generation.
The voice component (VTS) is built on Qwen3-Instruct \cite{qwen3technicalreport}.
All experiments are run on NVIDIA H200 GPUs. Details (prompts, iteration budget, and hyperparameters) are provided in the Supp. 2.

\noindent\textbf{Baselines.}
We compare MUSE with representative storytelling agents, including Vlogger \cite{Zhuang_2024_CVPR}, AnimDirector \cite{10.1145/3680528.3687688}, MMStoryAgent \cite{hu2024storyagent}, V-GOT \cite{2412.02259}, and MovieAgent \cite{wu2025automated}.
When a baseline does not support a modality (e.g., audio narration), we evaluate it on the supported outputs only and mark missing dimensions accordingly.
\input{tables/musebench}

\noindent\textbf{Metrics.}
On ViStoryBench, we report identity and style consistency using Character Identity Score (CIDS) and Character Style Distance (CSD) in both Cross- and Self-mode; prompt adherence using Prompt Alignment (PA), instance-level alignment (IA), and character count matching (CM); and perceptual quality and diversity using Inception Score (Inc), Aesthetic Predictor (Aes), and Copy-Paste (CP).
On MUSEBench, we adopt visual metrics from ViStoryBench and additionally evaluate script quality, narration–visual alignment, and audio generation. Script metrics include SER, NSR, and CES; narration–visual alignment is assessed by Atmos., Synergy, and Grounding; and audio quality is measured by Age, Emotion, Prosody, and Clarity. Detailed definitions are provided in the Supp. 4.

\subsection{Quantitative Evaluation on ViStoryBench}

\input{tables/visbench}
We first evaluate visual consistency on ViStoryBench, which provides reference character images and focuses on visual coherence.
Table~\ref{tab:vistory_results} reports results on identity/style consistency, prompt alignment, and perceptual quality and diversity. MUSE consistently improves identity-related metrics (Cross-CIDS/Cross-CSD), indicating reduced character drift over long sequences, while maintaining competitive visual quality and diversity (Inc/Aes) with low copy-paste behavior (CP).
As ViStoryBench is reference-dependent and evaluates only visual outputs, it does not assess narrative-level planning or audio–visual consistency. Moreover, due to the planning module’s prompt rewriting, prompt-alignment metrics exhibit a moderate decline. We therefore further evaluate holistic storytelling performance on MUSEBench.
\subsection{Holistic Evaluation on MUSEBench}
As shown in Table~\ref{tab:musebench_holistic}, MUSE generates more narrative-driven expanded scripts (achieving better script-related scores). Under end-to-end testing, some methods lose certain identity anchors required for metric evaluation—their corresponding scores are marked with '-', whereas MUSE still attains competitive visual metrics. Additionally, it performs well in narration-visual consistency. Notably, we reproduced the narration component of MMStoryAgent using CosyVoice2: while MMStoryAgent’s audio generation exhibits good clarity and emotional expression, it lacks support for customized generation. In contrast, MUSE enables text-driven customized audio synthesis, though its quality is slightly inferior. We emphasize that MUSEBench is designed for comparative evaluation rather than absolute quality estimation, and we report all results alongside human studies to mitigate potential evaluator bias.
Additionally, we visualize generated sequences on MUSEBench in Fig.~\ref{fig:mainfigure}, selecting the first four shots produced by each method. It can be observed that MUSE’s outputs exhibit more diverse camera movements and a consistent overall style. MUSE better preserves character identity (including non-human characters) across shots and produces more diverse yet script-consistent compositions compared with baselines, which often exhibit structural drift or repetitive framing. The full video and more visual results are provided in the Supp. 5.
\input{tables/abalation}
\begin{figure}[t]
    \centering
    \includegraphics[width=0.99\linewidth]{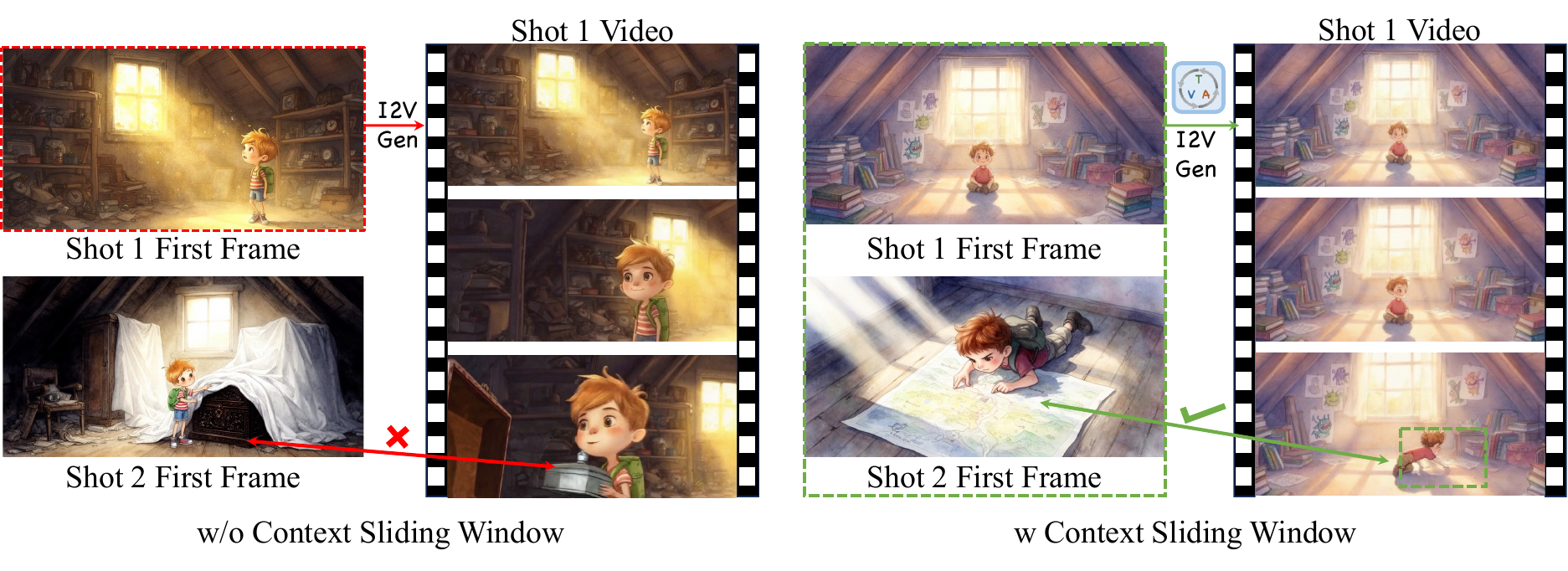}
    \vspace{-2mm}
    \caption{Impact of Context Sliding Window. Without temporal context, the character's action is discontinuous. With the window, MUSE maintains pose and object consistency across shots.}
    \label{fig:sliding_window}
    \vspace{-2mm}
\end{figure}
\subsection{Ablation Study \& In-depth Analysis}

\noindent\textbf{Closed-Loop Orchestration.} We conduct ablation experiments on MUSEBench with four experimental configurations: (1) Basemodel: direct feed-forward generation; (2) w/ Planning: Basemodel augmented solely with the planning module; (3) w/ Feedback: Basemodel with the feedback module; and (4) Full Mode. Since Closed-Loop Orchestration primarily acts on character audio generation in the audio component, it mainly impacts the Age and Prosody metrics among audio indicators, with the corresponding scores as follows: (1) 2.79/1.54; (2) 2.81/1.60; (3) 2.97/1.70; (4) 3.05/1.75. Specifically, the feed-forward component is primarily responsible for information integration, while the feedback component performs targeted revisions, resulting in more significant improvements. Additionally, as shown in the table, we focus on reporting visual-related metrics, and the results demonstrate the critical role of rational planning in storytelling systems. In contrast, the performance improvements of the feedback module are mainly concentrated on identity consistency preservation and style coherence. This phenomenon stems from the task-specific design of the feedback mechanism—it can provide targeted revision suggestions tailored to specific scenarios within the framework.

\noindent\textbf{Temporal Consistency via Context Sliding Window.}
Long-form video generation often suffers from state resets between clips.
Fig.~\ref{fig:sliding_window} shows that without the context sliding window, the character's interaction state is frequently broken across consecutive shots.
Conditioning each chunk on the terminal state of the previous one improves motion/state continuity and preserves object interactions.

\begin{figure}[t]
    \centering
    \includegraphics[width=0.99\linewidth]{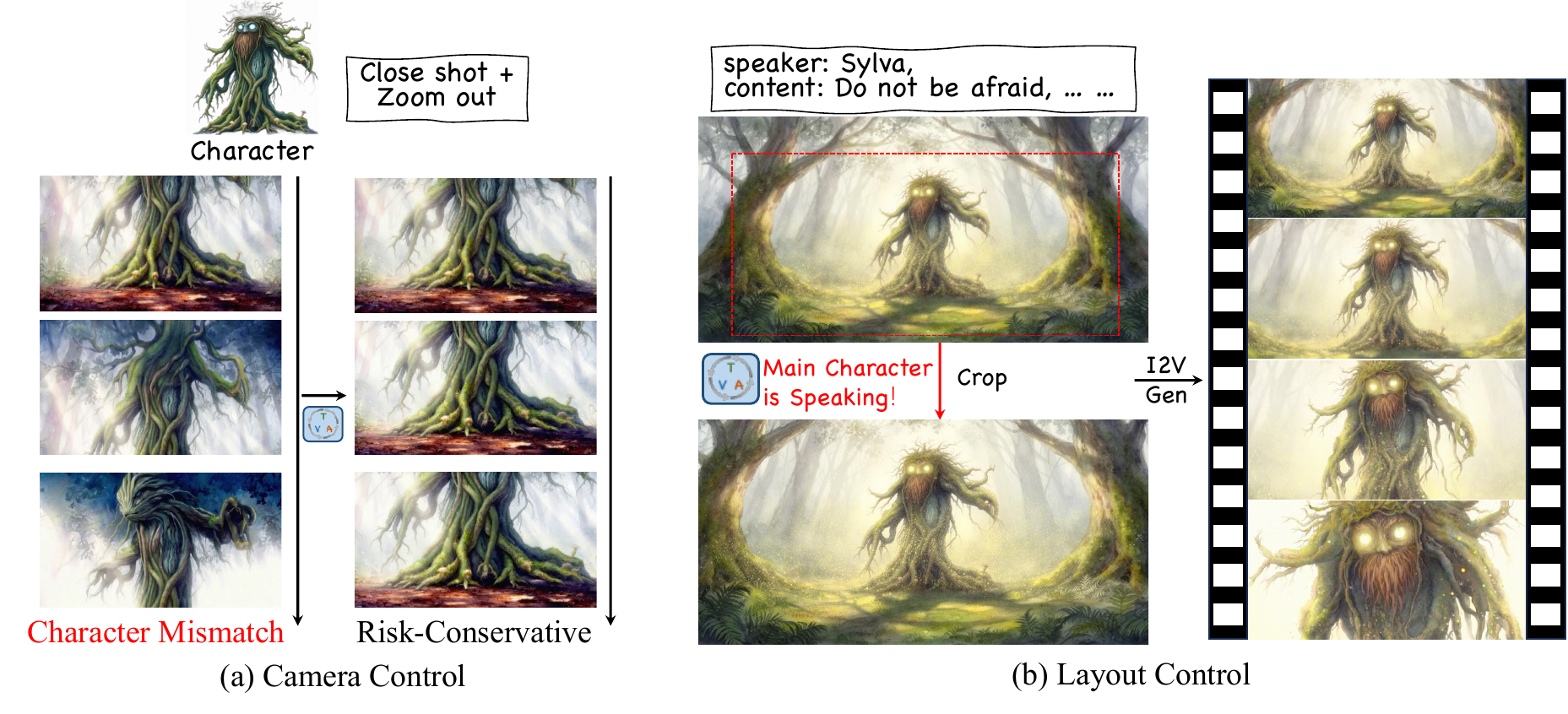} 
    \vspace{-2mm}
    \caption{Defensive Camera Control: MUSE adopts adaptive strategies across diverse scenarios to mitigate identity leakage.}
    \label{fig:camera_control}
    \vspace{-2mm}
\end{figure}
\noindent\textbf{Spatial Robustness via Defensive Camera Control.}
Complex scenes with dense backgrounds can amplify diffusion failures such as identity leakage. As illustrated in Fig.~\ref{fig:camera_control}, MUSE switches to a more conservative strategy when risk is high: for close-up shots, it refrains from using zoom-out camera movements within the current shot to prevent identity leakage caused by unintended character intrusion. Additionally, to enhance visual diversity, when characters in the frame are speaking, cropping is centered on the speaking characters.

\begin{figure}[t]
    \centering
    \includegraphics[width=0.99\linewidth]{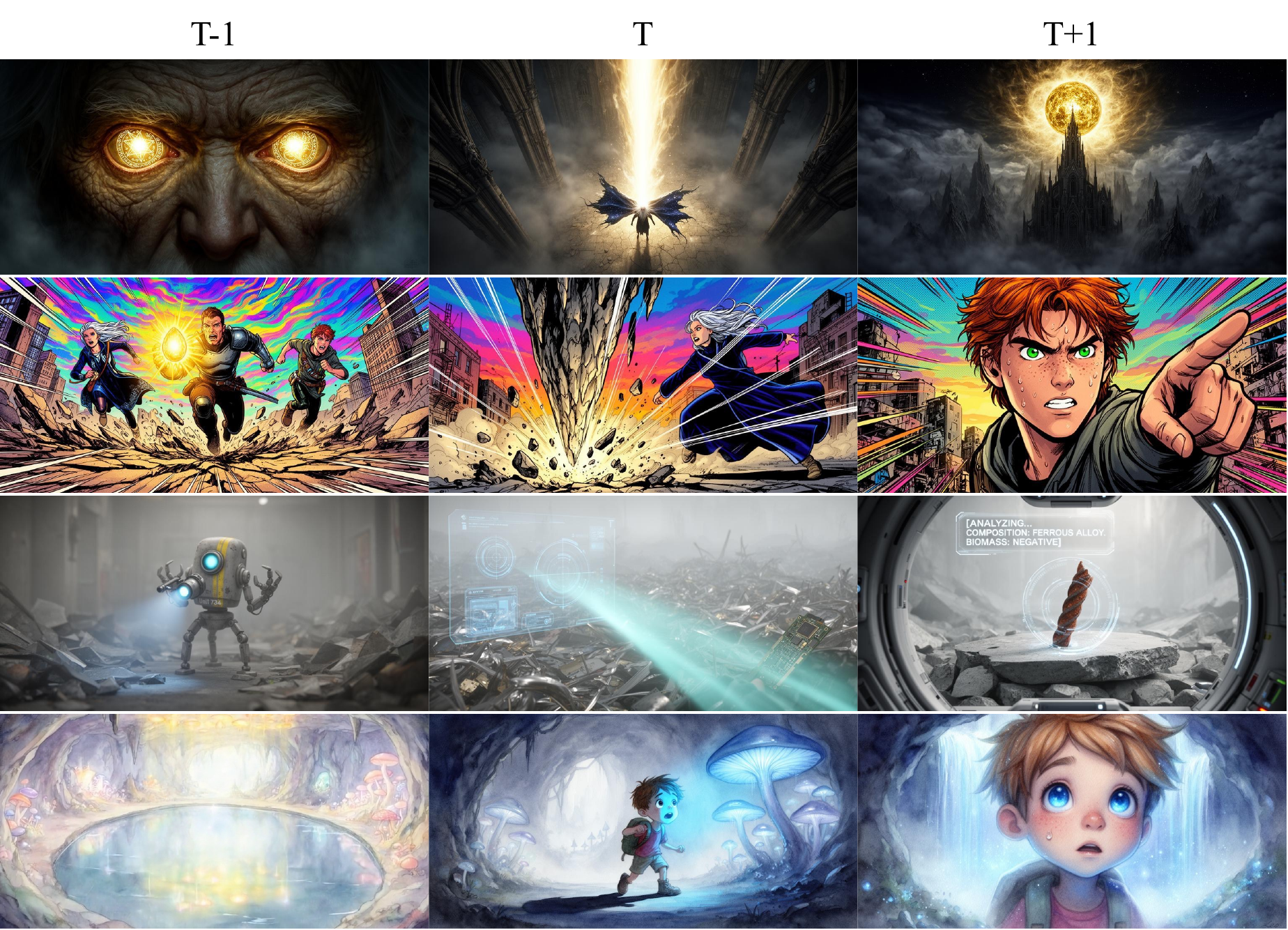} 
    \vspace{-2mm}
    \caption{Additional Visual Results: Selected consecutive shots from MUSE-generated videos.}
    \vspace{-2mm}
    \label{fig:morevis}
\end{figure}
\noindent\textbf{More Visual Results.} We present additional visual outputs generated by MUSE, selecting three consecutive scenes to demonstrate the diversity of its results. Thanks to the unified closed-loop execution, MUSE can devise diverse camera movement variations based on narrative progression and generate scene-specific visuals tailored to the current camera dynamics. MUSE not only produces videos across diverse styles but also exhibits notable advantages in scene variety and narrative continuity.

\begin{figure}[t]
    \centering
    \vspace{-2mm}
    \includegraphics[width=0.99\linewidth]{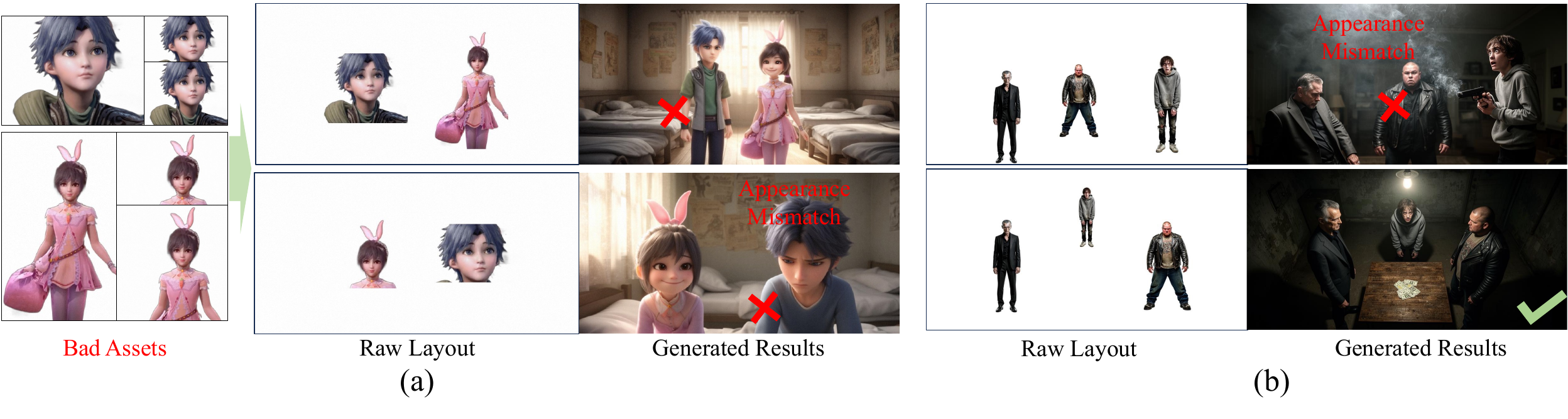} 
    \caption{Failure Cases: Appearance Mismatch Induced by Layout-Guided Generation.}
    \label{fig:failure}
    \vspace{-2mm}
\end{figure}

\noindent\textbf{Failure Cases.} Despite MUSE’s capability to integrate multimodal assets for generating consistent videos, it may produce outputs with inconsistencies under complex scenarios or when character assets are suboptimal. For instance, as shown in Figure~\ref{fig:failure} (a), MUSE requires full-body character representations to construct assets adaptable to diverse camera movements. However, the abundance of half-body character images in ViStoryBench leads to considerable identity discrepancies across varying camera dynamics. As illustrated in Figure~\ref{fig:failure} (b), in complex scenes involving multiple characters, preserving individual character information while generating natural movements remains a challenging problem.

\noindent\textbf{Human--Metric Alignment.}
To assess the reliability of MUSEBench, we measure the Pearson correlation between automatic scores and human ratings on a subset of 140 samples.
As reported in Table~\ref{tab:human_alignment}, MUSEBench aligns with human judgments on visual and narrative dimensions, while audio exhibits moderate correlation due to higher subjectivity.
\input{tables/humanalign}

%% file: tables/musebench.tex
\begin{table*}[t]
\centering
\caption{Holistic Evaluation on MUSEBench.
Holistic evaluation results across script quality, visual consistency, visual–script alignment, and audio generation.}
\label{tab:musebench_holistic}
\resizebox{\linewidth}{!}{
\begin{tabular}{l|ccc|ccccccc|ccccccc|cccc}
\toprule
\multirow{2}{*}{\textbf{Method}} 
& \multicolumn{3}{c|}{\textbf{Scripts}} 
& \multicolumn{7}{c|}{\textbf{Visual}} 
& \multicolumn{7}{c|}{\textbf{Visual--Script}} 
& \multicolumn{4}{c}{\textbf{Audio}} \\
\cmidrule(lr){2-4}\cmidrule(lr){5-11}\cmidrule(lr){12-18}\cmidrule(lr){19-22}
& \textbf{NSR$\uparrow$} 
& \textbf{SER$\uparrow$} 
& \textbf{CES$\uparrow$}
& \textbf{CIDS-C$\uparrow$} 
& \textbf{CIDS-S$\uparrow$} 
& \textbf{CSD-S$\uparrow$} 
& \textbf{CSD-C$\uparrow$} 
& \textbf{CP$\downarrow$} 
& \textbf{Inc$\uparrow$} 
& \textbf{OCCM$\uparrow$}
& \textbf{Scene$\uparrow$} 
& \textbf{CA$\uparrow$} 
& \textbf{Camera$\uparrow$} 
& \textbf{Atmos.$\uparrow$} 
& \textbf{Synergy$\uparrow$} 
& \textbf{Nes$\uparrow$} 
& \textbf{Grounding$\uparrow$}
& \textbf{Age$\uparrow$} 
& \textbf{Emotion$\uparrow$} 
& \textbf{Prosody$\uparrow$} 
& \textbf{Clarity$\uparrow$} \\
\midrule
AnimDirector \cite{10.1145/3680528.3687688}
& \cellcolor{cyan!10}3.53 
& 1.73 
& 1.87 
& - & - 
& 0.638 
& - 
& \cellcolor{cyan!10}0.276 
& 3.53 
& \cellcolor{cyan!20}\textbf{85.3}
& \cellcolor{cyan!20}\textbf{3.60} 
& \cellcolor{cyan!10}3.11 
& - & - & - & - & -
& - & - & - & - \\
MMStoryAgent* \cite{hu2024storyagent}
& 2.89 
& 1.59 
& 2.28 
& - & - 
& \cellcolor{cyan!20}\textbf{0.791} 
& - 
& - 
& 2.31 
& 27.7
& 2.09 
& 1.50 
& - 
& \cellcolor{cyan!10}2.90 
& \cellcolor{cyan!10}2.35 
& \cellcolor{cyan!10}2.45 
& \cellcolor{cyan!10}0.428
& - 
& \cellcolor{cyan!10}\textbf{2.01} 
& - 
& \cellcolor{cyan!20}\textbf{4.38} \\
V-GOT \cite{2412.02259}
& 2.23 
& 1.97 
& 2.2 
& - & - 
& \cellcolor{cyan!20}\textbf{0.791} 
& \cellcolor{cyan!20}\textbf{0.783} 
& - 
& 3.31 
& 66.9
& 2.63 
& 1.82 
& \cellcolor{cyan!20}\textbf{2.63} 
& - & - & - & -
& - & - & - & - \\
MovieAgent \cite{wu2025automated}
& 3.50 
& \cellcolor{cyan!10}2.53 
& \cellcolor{cyan!10}3.73 
& - & - 
& 0.624 
& 0.317 
& - 
& \cellcolor{cyan!10}4.87 
& 74.3
& 2.48 
& 2.07 
& 2.26 
& - & - & - & -
& - & - & - & - \\
\textbf{MUSE (Ours)} 
& \cellcolor{cyan!20}\textbf{3.70} 
& \cellcolor{cyan!20}\textbf{3.67} 
& \cellcolor{cyan!20}\textbf{3.93} 
& \cellcolor{cyan!20}\textbf{0.714} 
& \cellcolor{cyan!20}\textbf{0.712} 
& \cellcolor{cyan!10}0.710 
& \cellcolor{cyan!10}0.637 
& \cellcolor{cyan!20}\textbf{0.158} 
& \cellcolor{cyan!20}\textbf{4.95} 
& \cellcolor{cyan!10}81.2
& \cellcolor{cyan!10}3.58 
& \cellcolor{cyan!20}\textbf{3.40} 
& \cellcolor{cyan!10}2.60 
& \cellcolor{cyan!10}2.50 
& \cellcolor{cyan!10}2.82 
& \cellcolor{cyan!10}3.17 
& \cellcolor{cyan!20}\textbf{0.857}
& \cellcolor{cyan!20}\textbf{3.05} 
& \cellcolor{cyan!10}1.57 
& \cellcolor{cyan!20}\textbf{1.75} 
& \cellcolor{cyan!10}4.17 \\
\bottomrule
\end{tabular}
}
\end{table*}

%% file: tables/visbench.tex
\begin{table}[t]
    \centering
    \caption{Quantitative Comparison on ViStoryBench.
MUSE achieves strong performance on identity-related measures, particularly on Cross-mode metrics that evaluate fidelity to the initial character profile. Best and second-best results are highlighted in darker and lighter colors, respectively.}
    \label{tab:vistory_results}
    \resizebox{\linewidth}{!}{
    \begin{tabular}{l|cc|cc|cc|cccc}
        \toprule
        \multirow{2}{*}{\textbf{Method}} 
        & \multicolumn{2}{c|}{\textbf{CSD} $\uparrow$} 
        & \multicolumn{2}{c|}{\textbf{CIDS} $\uparrow$} 
        & \multicolumn{2}{c|}{\textbf{PA} $\uparrow$} 
        & \multirow{2}{*}{\textbf{CM$\uparrow$}} 
        & \multirow{2}{*}{\textbf{Inc$\uparrow$}} 
        & \multirow{2}{*}{\textbf{Aes$\uparrow$}} 
        & \multirow{2}{*}{\textbf{CP$\downarrow$}} \\
        & Cross & Self & Cross & Self & Scene & IA & & & & \\
        \midrule
        Vlogger \cite{Zhuang_2024_CVPR} 
        & 0.259 & 0.453 
        & 0.362 & 0.554 
        & 0.171 & 2.44 
        & \cellcolor{cyan!20}\textbf{76.6} 
        & 9.77 
        & 4.28 
        & 0.200 \\
        
        AnimDirector \cite{10.1145/3680528.3687688}
        & 0.288 & 0.510 
        & \cellcolor{cyan!10}0.401 & \cellcolor{cyan!10}0.578 
        & \cellcolor{cyan!20}\textbf{3.64} & \cellcolor{cyan!10}2.69 
        & \cellcolor{cyan!10}67.4 
        & 12.02 
        & 5.59 
        & 0.212 \\
        
        MMStoryAgent \cite{hu2024storyagent}
        & 0.238 & \cellcolor{cyan!20}\textbf{0.669} 
        & 0.388 & \cellcolor{cyan!20}\textbf{0.596} 
        & 2.92 & 1.63 
        & 61.5 
        & 9.09 
        & \cellcolor{cyan!10}5.88 
        & 0.198 \\
        
        V-GOT \cite{2412.02259}
        & 0.232 & 0.606 
        & 0.322 & 0.567 
        & 1.11 & 0.78 
        & 65.2 
        & 13.02 
        & \cellcolor{cyan!20}\textbf{6.16} 
        & \cellcolor{cyan!20}\textbf{0.171} \\
        
        MovieAgent \cite{wu2025automated}
        & \cellcolor{cyan!10}0.299 & 0.479 
        & 0.400 & 0.544 
        & \cellcolor{cyan!10}3.50 & \cellcolor{cyan!20}\textbf{2.73} 
        & 64.6 
        & \cellcolor{cyan!20}\textbf{14.99} 
        & 5.32 
        & 0.209 \\
        
        \textbf{MUSE (Ours)} 
        & \cellcolor{cyan!20}\textbf{0.412} & \cellcolor{cyan!10}0.614 
        & \cellcolor{cyan!20}\textbf{0.453} & 0.548 
        & 3.17 & 2.17 
        & 63.5 
        & \cellcolor{cyan!10}13.68 
        & 5.38 
        & \cellcolor{cyan!10}0.192 \\
        \bottomrule
    \end{tabular}
    }
\end{table}

%% file: tables/abalation.tex
\begin{table}[t]
    \centering
    \caption{Ablation Results on MUSEBench.
Ablation study evaluating the impact of planning and feedback components on holistic storytelling performance across visual, visual–script related metrics.}
    \label{tab:ablation_musebench}
    \vspace{-2mm}
    \resizebox{\linewidth}{!}{
    \begin{tabular}{l|ccccccc|ccccccc}
        \toprule
        \multirow{2}{*}{\textbf{Setting}} 
        & \multicolumn{7}{c|}{\textbf{Visual}}                                                                                  & \multicolumn{7}{c}{\textbf{Visual-Script}}                                                                                \\
        & \textbf{CIDS-C} $\uparrow$ & \textbf{CIDS-S} $\uparrow$ & \textbf{CSD-S} $\uparrow$ & \textbf{CSD-C} $\uparrow$ & \textbf{CP} $\downarrow$ & \textbf{Inc} $\uparrow$ & \textbf{OCCM} $\uparrow$ & \textbf{Scene} $\uparrow$ & \textbf{CA} $\uparrow$ & \textbf{Camera} $\uparrow$ & \textbf{Atmos.} $\uparrow$ & \textbf{Synergy} $\uparrow$ & \textbf{Nes} $\uparrow$ & \textbf{Grounding} $\uparrow$ \\
        \midrule
        Basemodel                  & 0.609 & 0.597 & 0.601 & 0.372 & 0.227 & 4.79 & 74.7 & 3.17 & 3.27 & 2.43 & 2.41 & 2.47 & 2.64 & 0.619 \\
        
        w/ Planning                & \cellcolor{cyan!10}0.697 & 0.671 & 0.701 & 0.617 & \cellcolor{cyan!10}0.173 & \cellcolor{cyan!20}\textbf{4.97} & \cellcolor{cyan!10}80.17 & \cellcolor{cyan!10}3.51 & 3.29 & \cellcolor{cyan!20}\textbf{2.61} & 2.47 & \cellcolor{cyan!10}2.79 & \cellcolor{cyan!10}3.04 & \cellcolor{cyan!10}0.791 \\
        
        w/ Feedback                & 0.671 & \cellcolor{cyan!10}0.691 & \cellcolor{cyan!10}0.703 & \cellcolor{cyan!10}0.623 & 0.187 & 4.91 & 77.3 & 3.37 & 3.37 & 2.47 & \cellcolor{cyan!10}2.49 & 2.53 & 2.71 & 0.637 \\
        
        \textbf{Full Mode}       & \cellcolor{cyan!20}\textbf{0.714} & \cellcolor{cyan!20}\textbf{0.712} & \cellcolor{cyan!20}\textbf{0.710} & \cellcolor{cyan!20}\textbf{0.637} & \cellcolor{cyan!20}\textbf{0.158} & \cellcolor{cyan!10}4.95 & \cellcolor{cyan!20}\textbf{81.2} & \cellcolor{cyan!20}\textbf{3.58} & \cellcolor{cyan!20}\textbf{3.40} & \cellcolor{cyan!10}2.60 & \cellcolor{cyan!20}\textbf{2.50} & \cellcolor{cyan!20}\textbf{2.82} & \cellcolor{cyan!20}\textbf{3.17} & \cellcolor{cyan!20}\textbf{0.857} \\
        \bottomrule
    \end{tabular}
    \vspace{-4mm}
    }
\end{table}

%% file: tables/humanalign.tex
\begin{table}[t]
    \centering
    \caption{\textbf{Human-Agent Alignment.} We report the Pearson correlation coefficient ($r$) between MUSEBench automated scores and human expert ratings across three modalities.}
    \label{tab:human_alignment}
    \resizebox{0.9\linewidth}{!}{
    \begin{tabular}{l|c c c}
        \toprule
        \textbf{Modality} & \textbf{Scripts} & \textbf{Visual} & \textbf{Audio} \\
        \midrule
        \textit{Metric Focus} & \small{Logical Consistency} & \small{Visual Fidelity} & \small{Timbre/Quality} \\
        Pearson ($r$) & 0.64 & 0.74 & 0.49 \\
        \bottomrule
    \end{tabular}
    }
\end{table}
\vspace{-2mm}

%% file: sections/conclusion.tex
\section{Conclusion and Future Work}
\label{sec:conclusion}
\noindent\textbf{Conclusion.}
We presented MUSE, a multi-agent framework for long-form audio-visual storytelling that addresses the intent–execution gap between high-level narrative prompts and reliable shot-level generation. By reformulating storytelling as a closed-loop constraint enforcement process, MUSE enables explicit planning, verification, and targeted revision across vision, audio, and time. Extensive experiments and ablations demonstrate that enforcing global narrative constraints substantially improves long-horizon coherence and cross-modal consistency in open-ended storytelling.

\noindent\textbf{Future Work.}
Several challenges remain for long-form audio-visual storytelling. First, maintaining stable character identities in crowded or occluded scenes remains difficult, especially under frequent viewpoint changes. Second, while current systems can anchor coarse vocal traits, richer control over expressive speech and emotional dynamics is needed for more natural dialogue. 

%% file: sections/supp.tex

\section{Reproducibility Checklist}
\label{app:reproducibility}

To ensure the reproducibility of MUSE, we provide the following resources and specifications. The maximum number of iterations is set to 5. If all attempts fail after 5 iterations, the result with the highest score is selected:
\begin{itemize}
    \item Code Availability: The complete source code, including the multi-agent orchestration framework, VTS inference scripts, and MUSEBench evaluation toolkit, is included in the supplementary zip file. A public GitHub repository will be released upon acceptance.
    \item Model Versions:
    \begin{itemize}
        \item \textit{Reasoning Backbone:} \texttt{Gemini-2.5-Pro}\cite{comanici2025gemini} (Temperature: 0.7 for planning, 0.2 for critique).
        \item \textit{Visual Backbone:} \texttt{Flux.2-Dev}\cite{flux-2-2025}(Guidance Scale: 3.5, Steps: 28) and \texttt{Wan2.2-I2V-A14B}\cite{2503.20314}.
        \item \textit{Audio Backbone:} Custom VTS-model (See Section \ref{app:vts_details} for training specifics).
    \end{itemize}
    \item Data:
    \begin{itemize}
        \item MUSEBench: Comprising 30 structured narrative prompts and multi-dimensional evaluation rubrics.
        \item ViStoryBench\cite{zhuang2025vistorybench}: Comprising 80 stories adapted for long-form consistency evaluation.
    \end{itemize}
\end{itemize}
Furthermore, we need to emphasize that MUSE is a control-and-verification layer, not tied to any specific generative models such as Flux or Wan.

\section{VTS Model Implementation Details}
\label{app:vts_details}

To enable Zero-Shot Identity Anchoring, we developed a bespoke Vocal Trait Synthesis (VTS) model. Unlike generic TTS APIs, VTS is optimized for semantic-to-acoustic projection.

\paragraph{Data Construction.} 
We curated a large-scale proprietary dataset containing 20,000 hours of high-fidelity audio. To construct semantic-acoustic pairs, we utilized \texttt{Gemini-2.5-Pro} to annotate each audio clip with fine-grained conditional metadata, including \textit{Gender}, \textit{Age}, \textit{Timbre} (e.g., raspy, bright), and \textit{Prosody} (e.g., whispering, shouting).

\paragraph{Architecture \& Tokenization.}
The VTS model is built upon the Qwen3-1.7B\cite{qwen3technicalreport} architecture, adapted for cross-modal generation:
\begin{itemize}
    \item Audio Encoder: We employ a custom \texttt{XY-Tokenizer} (a discrete neural audio codec) to quantize audio waveforms into discrete acoustic tokens.
    \item Condition Encoder: Semantic conditions are processed via the Qwen3 text encoder.
    \item Input Formatting: The model is trained as a decoder-only transformer following the pattern: $\texttt{[Conditions] + [Acoustic Tokens]}$.
\end{itemize}

\paragraph{Training Strategy.}
We adopted a two-stage training paradigm to ensure robust zero-shot generalization:
\begin{enumerate}
    \item Pre-training: The model was first pre-trained on 100,000 hours of unconditioned audio data to learn robust acoustic modeling and phoneme alignment.
    \item Instruction Tuning: We fine-tuned the model on the 20,000 hours of annotated data. This stage aligns the latent acoustic space with the semantic descriptors used in our Pre-production Agent.
\end{enumerate}

\paragraph{Zero-Shot Inference.}
Crucially, unlike traditional Voice Cloning which requires an external 3-second reference audio, VTS generates the reference audio \textit{itself} based solely on the script's character profile. Once synthesized during the Pre-production phase, this audio clip is frozen and serves as the ground truth anchor for all subsequent dialogue generation, strictly enforcing ontological consistency.
\paragraph{Comparisions.}To evaluate the audio continuation capability of TTS models, we selected 400 ground-truth (GT) audio clips. The textual descriptions corresponding to these GT audio clips were annotated using Gemini2.5 Pro. We then leveraged MUSEBench to assess the similarity between the original GT audio and the continuation-generated audio across multiple dimensions. The evaluation results are presented in Table \ref{tab:audio_continuation_similarity}.

\begin{table}[h]
    \centering
    \caption{Similarity Evaluation Results Across Audio Dimensions. Quantitative comparison of audio continuation performance across three TTS models (CosyVoice2, F5-TTS, and VTS) on MUSEBench's core audio metrics, where higher scores indicate better consistency with the original GT audio.}
    \label{tab:audio_continuation_similarity}
    \resizebox{\linewidth}{!}{
    \begin{tabular}{l|cccc}
        \toprule
        \textbf{Model} 
        & \textbf{Age$\uparrow$} 
        & \textbf{Emotion$\uparrow$} 
        & \textbf{Prosody$\uparrow$} 
        & \textbf{Clarity$\uparrow$} \\
        \midrule
        CosyVoice2\cite{Du2024-ze} 
        & 3.7 
        & 2.02 
        & 2.08 
        & 4.38 \\
        
        F5-TTS\cite{chen-etal-2025-f5} 
        & \cellcolor{cyan!10}3.76 
        & \cellcolor{cyan!10}2.1 
        & \cellcolor{cyan!10}2.2 
        & \cellcolor{cyan!20}\textbf{4.92} \\
        
        VTS (Ours)
        & \cellcolor{cyan!20}\textbf{3.97} 
        & \cellcolor{cyan!20}\textbf{2.23} 
        & \cellcolor{cyan!20}\textbf{2.44} 
        & \cellcolor{cyan!10}4.59 \\
        \bottomrule
    \end{tabular}
    }
\end{table}
Table \ref{tab:audio_continuation_similarity} presents MUSEBench evaluation results of three TTS models (CosyVoice2, F5-TTS, VTS (Ours)) on audio continuation across four core dimensions (Age, Emotion, Prosody, Clarity). Key findings:
Age Consistency: VTS achieves the highest score (3.97), outperforming CosyVoice2 (3.7) and F5-TTS (3.76), effectively preserving age-related vocal traits for long-form narrative identity consistency.
Emotion \& Prosody: VTS leads with 2.23 (Emotion) and 2.44 (Prosody), validating its semantic-driven design—deriving vocal anchors from textual descriptors to maintain coherent emotional tone and rhythm, critical for natural storytelling.
Clarity: F5-TTS tops with 4.92, while VTS (4.59) maintains competitiveness, prioritizing narrative-critical dimensions over pure clarity.
Overall: VTS delivers superior narrative-aligned audio continuation, excelling in cross-modal identity consistency and immersion—aligning with MUSE’s core goal of global constraint enforcement via closed-loop orchestration.

\section{Full Agent Interface Specification}
\label{app:agent_interface}
To ensure the reproducibility and rigorous definition of our agentic workflow, we provide the formal specification of the inter-agent communication protocols.

\noindent\textbf{Overview of Agent Roles}
Table \ref{tab:agent_overview} provides a compact mapping between the mathematical formulations in Section 3 and the specific engineering modules.
\input{tables/agentoverview}

To ensure reproducibility, we formally define the communication protocols for the Pre-production, Production, and Post-production phases. These schemas govern the transformation from abstract intent to executable constraints.

\noindent\textbf{Phase 1: Pre-production Team (Identity Anchoring)}
The Pre-production phase initializes immutable identity constraints through a multi-stage generation and verification process.

\noindent{I. Context Injection: Story Style Profile.}
The \textit{Style Analyzer} locks the global aesthetic (Phase 0 in implementation) to prevent style drift.
\begin{lstlisting}[language=json]
{
  "context_type": "global_style_injection",
  "analyzed_profile": {
    "genre": "Slice of Life / Urban Drama",
    "tone": "Melancholic, introspective",
    "art_style": "Watercolor Storybook",
    // These keywords are prefixed to EVERY character prompt
    "style_modifier": "watercolor illustration, soft edges, ink and wash",
    "scene_guide": "watercolor landscape painting, wet-on-wet technique"
  }
}
\end{lstlisting}

\noindent{II. Forward Stream ($\Phi_{pre}$): Multimodal Synthesis.}
The generator produces both visual and audio assets in parallel.
\begin{lstlisting}[language=json]
{
  "task": "generate_character_assets",
  "character_id": "Arthur",
  // Visual Generation Params
  "visual_prompt": {
    "prefix": "FULL BODY PORTRAIT, Single Character",
    "style_anchor": "watercolor illustration, soft edges...", 
    "appearance": "38yo male, pale, dark circles, slumped shoulders",
    "constraint": "head-to-toe, feet visible, simple white background"
  },
  // Audio Generation Params (VTS)
  "audio_prompt": {
    "acoustic_features": "Mid-to-low pitch, lack of chest resonance, flat tone",
    "rhythmic_features": "Slow, monotonous, sigh-heavy",
    "target_transcript": "I need to find the Whispering Giant..."
  }
}
\end{lstlisting}

\noindent{III. Backward Stream ($\Psi_{pre}$): Hierarchical Critique.}
The Critic Agent performs verification at two levels: \textit{Atomic Asset Audit} and \textit{Global Consistency Audit}.

\noindent{Level 1: Atomic Asset Audit (Image \& Audio).
\begin{lstlisting}[language=json]
{
  "audit_level": "atomic_asset",
  // 1. Visual Evaluation (derived from ImageEvaluation Class)
  "visual_critique": {
    "framing_check": {
      "is_full_body": true,       // Critical Gate
      "feet_visible": true,
      "head_to_toe_in_frame": true
    },
    "anatomical_integrity": {
      "score": 10,
      "hands_and_fingers": "normal",  // Checks for extra digits
      "face_structure": "normal"      // Checks for melted faces
    }
  },
  // 2. Audio Evaluation (derived from AudioEvaluation Class)
  "audio_critique": {
    "voice_match": {
      "gender_match": true,
      "age_match": true,
      "timbre_match": "high"  // Does it sound like the description?
    },
    "performance_quality": {
      "emotion_accuracy": "low", // Error: sounded happy instead of tired
      "naturalness": 7
    },
    "audio_image_consistency": "low" // Cross-modal check: Voice vs Face
  }
}
\end{lstlisting}

\noindent{Level 2: Global Consistency Audit.}
Triggered only when multiple characters are generated.
\begin{lstlisting}[language=json]
{
  "audit_level": "cross_character_consistency",
  "input_batch": ["image_arthur_final", "image_narrator_final"],
  "evaluation_metrics": {
    "visual_style_consistency": "high", // Do they look like the same movie?
    "script_style_match": "high",       // Do they match the 'Watercolor' genre?
    "detected_style": "watercolor illustration"
  },
  "overall_consistency_score": 9.5
}
\end{lstlisting}

\noindent{IV. Optimization Policy ($\Omega$): Adaptive Correction.}
Based on the error topology, the optimizer selects a specific repair strategy.
\begin{lstlisting}[language=json]
{
  "optimization_trigger": "AUDIO_FAILURE (Score 6.7/10)",
  "error_diagnosis": {
    "modality": "audio",
    "issue": "EMOTION_MISMATCH",
    "details": ["Voice sounds too energetic", "Lacks vocal fry"]
  },
  "selected_strategy": "REWRITE_PROMPT", 
  "execution_plan": {
    "action": "refine_audio_descriptor",
    "reasoning": "LLM adds 'gravelly vocal fry' and 'exhausted sighs' to prompt.",
    "new_descriptor": {
      "Acoustic": "Pronounced gravelly vocal fry at end of phrases...",
      "Rhythmic": "Slower, pause-heavy"
    }
  }
}
\end{lstlisting}

\noindent\textbf{Phase 2: Production Team (Spatial Composition)}
The Production phase ensures spatial fidelity through a three-step workflow: \textit{Dynamic Routing}, \textit{Geometric Layout Refinement}, and \textit{Feedback-Driven Synthesis}.

\noindent{I. Forward Stream ($\Phi_{prod}$): Layout Planning \& Guardrails.}
Before generation, the system performs intelligent routing and rigorously validates spatial constraints to prevent physical impossibilities.

\noindent{Step 1: Dynamic Routing (VLM Decision).}
The \textit{Prompt Translator} determines the rendering mode based on narrative focus.
\begin{lstlisting}[language=json]
{
  "task": "determine_layout_mode",
  "input": "Close-up of Arthur's hand gripping the briefcase...",
  "decision_logic": {
    "stage_1_non_face": true, // Body part detected -> "none" (T2I)
    "stage_2_facial": false,
    "stage_3_default": false
  },
  "final_decision": "none" // Path A: Pure Text-to-Image
}
\end{lstlisting}

\noindent{Step 2: Geometric Guardrails (BBox Processing).}
If layout is required, the system enforces spatial logic \textit{before} pixel generation.
\begin{lstlisting}[language=json]
{
  "process": "layout_refinement",
  "input_layout": {
    "Arthur": [0.45, 0.5, 0.55, 0.9] // Width 0.1 (Too thin!)
  },
  "guardrail_actions": [
    {
      "action": "resize_bbox",
      "target": "Arthur",
      "reason": "width_below_threshold (0.1 < 0.15)",
      "adjustment": "expand_from_center -> [0.425, 0.5, 0.575, 0.9]"
    },
    {
      "action": "resolve_overlap",
      "target": ["Arthur", "Prop_Briefcase"],
      "adjustment": "shift_edges_to_remove_intersection"
    }
  ],
  "final_layout_status": "optimized"
}
\end{lstlisting}

\noindent{II. Backward Stream ($\Psi_{prod}$): Visual \& Spatial Audit.}
The Critic Agent evaluates two distinct failure modes: \textit{Visual Artifacts} (e.g., sticker effect) and \textit{Spatial Conflicts} (e.g., illogical overlaps).

\begin{lstlisting}[language=json]
{
  "critique_type": "production_quality_check",
  
  // 1. Spatial Logic Check (Triggers 'Spatial Disentanglement')
  "spatial_logic_audit": {
    "bbox_overlap_detected": true, 
    "overlap_ratio": 0.15, // > 5% threshold
    "conflicting_subjects": ["Arthur", "Other_Char"],
    "physical_plausibility": false
  },

  // 2. Visual Integration Check (Triggers 'Guidance Modulation')
  "visual_integration": {
    "sticker_effect_severity": "Mild", 
    "shadow_logic": false, // Missing cast shadows
    "lighting_match": true
  },

  // 3. General Quality
  "overall_quality": {
    "aesthetic_score": 7.5,
    "limb_completeness": "Complete",
    "body_structure": "Reasonable"
  }
}
\end{lstlisting}

\noindent{III. Optimization Policy ($\Omega_{prod}$): Multimodal Correction.}
The policy engine dispatches distinct strategies based on the error topology: Guidance Modulation for visual artifacts and Spatial Disentanglement for layout conflicts.

\begin{lstlisting}[language=json]
[
  // Case A: Visual Artifact Correction (Flux Specific)
  {
    "optimization_trigger": "INTEGRATION_ISSUE",
    "error_diagnosis": {
      "issue": "STICKER_EFFECT_MILD",
      "description": "Subject edges are too sharp against background."
    },
    "selected_strategy": "GUIDANCE_MODULATION",
    "execution_plan": {
      "action": "adjust_flux_params",
      "param_update": {
        "guidance_scale": "3.5 -> 4.5", 
        "prompt_injection": "volumetric lighting, seamless blending"
      },
      "seed_update": "randomize"
    }
  },
  // Case B: Spatial Layout Correction (Geometric Guardrails)
  {
    "optimization_trigger": "SPATIAL_CONFLICT",
    "error_diagnosis": {
      "issue": "BBOX_OVERLAP_DETECTED",
      "description": "Character A and B overlap > 5% (Physical Collision)."
    },
    "selected_strategy": "SPATIAL_DISENTANGLEMENT",
    "execution_plan": {
      "action": "shift_coordinates",
      "layout_update": {
        "Arthur": "shift_left (x_max: 0.55 -> 0.52)",
        "Other_Char": "shift_right (x_min: 0.50 -> 0.53)"
      },
      "re-render": "regenerate_layout_canvas"
    }
  }
]
\end{lstlisting}

\noindent\textbf{Phase 3: Post-production Team (Temporal Synthesis)}
The Post-production phase manages the temporal dimension, employing strict boundary guards and visual stability checks to prevent narrative leakage and autoregressive degradation.

\noindent{I. Forward Stream ($\Phi_{post}$): Temporal Planning \& Framing.}
The \textit{Action Planner} prepares the visual context and generates progressive constraints before synthesis begins.

\noindent{Step 1: Intelligent Framing \& Camera Guardrails.}
To ensure character consistency, the system applies smart cropping and strictly limits camera movement for close-ups to prevent ``anatomical hallucination'' (e.g., zooming out from a face to reveal a distorted body).
\begin{lstlisting}[language=json]
{
  "task": "visual_context_preparation",
  "input_speaker": "Arthur",
  "framing_strategy": {
    "mode": "SMART_CROP_SPEAKER", // Detected speaker -> Crop to BBox
    "target_bbox": [0.3, 0.2, 0.7, 0.8],
    "resolution": "1024x1024"
  },
  "camera_guardrail": {
    "shot_type": "Close-up",
    "constraint_active": true,
    "forbidden_motion": ["Zoom Out", "Pull Back", "Wide Shot"],
    "enforced_motion": "Static or Slight Pan" // Prevent body hallucination
  }
}
\end{lstlisting}

\noindent{Step 2: Progressive Prompting with Boundary Awareness.}
The planner generates prompts for each chunk, explicitly defining what must happen (\texttt{current\_goal}) and what must NOT happen yet (\texttt{next\_scene\_forbidden}).
\begin{lstlisting}[language=json]
{
  "chunk_id": 1,
  "duration": 5,
  "narrative_focus": "Arthur reaches for the handle.",
  "boundary_guard": {
    "next_scene_event": "Door opens",
    "negative_prompt_injection": "door opening, seeing inside, open door"
  }
}
\end{lstlisting}

\noindent{II. Backward Stream ($\Psi_{post}$): Leakage \& Degradation Audit.}
The Critic performs two specific checks: ensuring the action hasn't progressed too far (Leakage) and checking for autoregressive visual decay (Over-exposure).

\begin{lstlisting}[language=json]
{
  "critique_type": "temporal_integrity_check",
  
  // 1. Narrative Leakage Analysis (Text vs Visual)
  "leakage_audit": {
    "current_chunk_text": "Arthur reaches for the handle",
    "next_chunk_text": "Door opens",
    "visual_analysis_end_frame": "Door is slightly ajar", // Detected Leakage
    "leakage_flag": true
  },

  // 2. Autoregressive Degradation Check (Last Frame Analysis)
  "visual_decay_audit": {
    "target": "final_chunk_last_frame",
    "histogram_analysis": {
      "highlight_clipping": 0.85, // > 0.8 indicates Over-exposure
      "contrast_collapse": true
    },
    "diagnosis": "BURN_OUT_DETECTED" // Common in long autoregressive chains
  }
}
\end{lstlisting}

\noindent{III. Optimization Policy ($\Omega_{post}$): Containment \& Replanning.}
The system employs distinct strategies for local boundary violations versus global quality degradation.

\begin{lstlisting}[language=json]
[
  // Strategy A: Narrative Containment (Fixing Leakage)
  {
    "trigger": "NARRATIVE_LEAKAGE",
    "action": "REGENERATE_CHUNK",
    "param_update": {
      "prompt_refinement": "emphasize 'hand touching handle only'",
      "negative_prompt_boost": "open door, interior view",
    }
  },
  // Strategy B: Global Replanning (Fixing Over-exposure)
  {
    "trigger": "VISUAL_BURN_OUT",
    "action": "REDUCE_AND_RESTART",
    "reasoning": "Too many recursion steps caused accumulation of high-frequency noise (burnout).",
    "execution_plan": {
      "new_segmentation": "Reduce chunk count (e.g., 3 chunks -> 2 chunks)",
      "denoising_adjustment": "Reduce strength in later chunks"
    }
  }
]
\end{lstlisting}

\section{MUSEBench Metric Specifications}
\label{app:musebench_metrics}

The evaluation metrics of MUSEBench cover visual, textual, audio, and cross-modal indicators, consisting of formula-based calculations and LLM-as-judge assessments. Several visual metrics are adopted from ViStoryBench with partial modifications.

\noindent\textbf{A. Narrative State Resolution (NSR)}$\uparrow$
\textbf{Definition:} NSR measures the ``Show, Don't Tell'' capability. It evaluates whether major state changes (e.g., emotional shifts, plot twists) are resolved through explicit character actions rather than merely stated in narration.

\noindent{Calculation Logic.}
The VLM identifies $N$ major state changes in the generated video/script. Each change $c_i$ is classified into a resolution level $L(c_i) \in \{0, 1, 2, 3\}$.
\begin{equation}
    NSR = \frac{\sum_{i=1}^{N} L(c_i)}{3 \times N} \times 100\%
\end{equation}
Where $L(c_i)$ is assigned as:
\begin{itemize}
    \item Fully Resolved (3 pts): Shows Setup + Action + Consequence.
    \item Partially Resolved (2 pts): Action or Consequence is shown; minor inference required.
    \item Weakly Resolved (1 pts): Outcome is stated; action is symbolic or minimal.
\end{itemize}

\noindent{Judge Prompt (Abbreviated).
\begin{lstlisting}[language=json]
{
  "task": "Evaluate Narrative State Resolution",
  "step_1": "Identify all major state changes (decisions, emotional shifts).",
  "step_2": "Classify resolution level based on visual evidence.",
  "rubric": {
    "Fully_Resolved": "Viewer sees HOW change occurred (no inference needed).",
    "Not_Resolved": "Change is declared in narration/dialogue but not shown."
  },
  "output_format": {
    "state_changes": [
      {
        "description": "Character A decides to trust B",
        "shots_involved": ["Shot 5", "Shot 6"],
        "classification": "FULLY_RESOLVED",
        "justification": "Shot 5 shows hesitation, Shot 6 shows handshake."
      }
    ],
    "resolution_rate": 0.85
  }
}
\end{lstlisting}

\noindent\textbf{B. Story Expansion Richness (SER)}$\uparrow$
\textbf{Definition:} SER quantifies the agent's ability to expand a brief user prompt into a multi-dimensional narrative. It combines quantitative expansion (shot count) with qualitative depth.

\noindent{Calculation Logic.}
SER is a composite score derived from five qualitative dimensions ($D_{qual}$) and a quantitative multiplier ($M_{quant}$).
\begin{equation}
    SER_{raw} = \left( \frac{1}{5} \sum_{k=1}^{5} Score(D_k) \right) \times M_{quant}(Shots)
\end{equation}
\begin{itemize}
    \item Qualitative Dimensions ($D_k$): 1. Character Depth; 2. World-Building; 3. Thematic Expansion; 4. Plot Complexity; 5. Emotional Tonal Range. (Each scored 0-5).
    \item Quantitative Multiplier ($M_{quant}$):
    \begin{itemize}
        \item $<15$ shots: $\times 0.6$
        \item $25-39$ shots: $\times 1.0$
        \item $\ge 55$ shots: $\times 1.2$ (Rewards long-form consistency)
    \end{itemize}
\end{itemize}

\noindent{Judge Prompt (Abbreviated).}
\begin{lstlisting}[language=json]
{
  "task": "Evaluate Story Expansion Richness",
  "input": {"user_prompt": "...", "generated_script": "..."},
  "dimensions": {
    "Character_Depth": "Do characters have internal conflicts/arcs?",
    "World_Building": "Is the setting historically/culturally rich?",
    "Thematic_Expansion": "Are there symbolic layers beyond the plot?"
  },
  "scoring_criteria": {
    "5_points": "Exceptional. Adds backstory, symbolism, and subplots.",
    "1_point": "Insufficient. Bare plot outline matching prompt exactly."
  }
}
\end{lstlisting}

\noindent\textbf{C. Creative Elaboration Score (CES)}$\uparrow$
\textbf{Definition:} CES measures the structural and psychological sophistication of the narrative. Unlike SER which focuses on ``expansion amount,'' CES focuses on ``artistic quality'' and ``narrative architecture.''

\noindent{Calculation Logic.}
The score determines the Creative Sophistication Index ($CSI$) based on five high-level narrative features.
\begin{equation}
    CES = \text{MapToLikert}\left( \frac{1}{5} \sum_{m=1}^{5} Sophistication(F_m) \right)
\end{equation}
Features ($F_m$) include:
\begin{enumerate}
    \item Narrative Architecture: Non-linear elements, parallel storylines, framing devices.
    \item Character Interiority: Representation of memories, fears, or psychological conflicts.
    \item Symbolic Design: Use of visual metaphors or recurring motifs.
    \item Meta-Cognitive Layer: Presence of directorial reasoning (Chain-of-Thought) in script.
    \item World Implication: Environmental details that imply a larger history.
\end{enumerate}

\noindent{Judge Prompt (Abbreviated).}
\begin{lstlisting}[language=json]
{
  "task": "Evaluate Creative Elaboration",
  "rubric": {
    "Narrative_Architecture": {
      "5": "Complex structure (flashbacks, parallel editing).",
      "1": "Random sequence of events."
    },
    "Character_Interiority": {
      "5": "Profound. Inner life drives action.",
      "1": "Archetypal. Characters are plot devices."
    }
  },
  "consistency_check": "If any dimension score is 5, Final Score cannot be < 3."
}
\end{lstlisting}

We adopt a comprehensive evaluation suite combining established metrics from ViStoryBench~\cite{zhuang2025vistorybench} with novel metrics designed for audio-visual narrative synergy.

\noindent\textbf{D. Identity and Consistency (Adapted)}
We employ the standard metrics defined in ViStoryBench but introduce a critical architectural modification to the Character Identity Score (CIDS).

\paragraph{CIDS-C/S (Modified via CLIP).}
Unlike the original CIDS which relies on face-recognition models (e.g., InsightFace), we propose a \textbf{CLIP-based CIDS} to support generalized character integrity (including costume and body shape), not just facial features.
\begin{itemize}
    \item \textbf{CIDS-C (Consistency with Reference) $\uparrow$:} We use GroundingDINO to detect character regions and compute the CLIP-ViT-L/14 feature cosine similarity between the generated character $x_{gen}$ and the anchor reference $x_{ref}$.
    \item \textbf{CIDS-S (Self-Consistency) $\uparrow$:} The pairwise cosine similarity average among all generated appearances of the same character within a story.
\end{itemize}

\paragraph{Standard Visual Metrics.}
We report the following metrics to ensure comparability, The calculation methods of these metrics are directly adopted from ViStoryBench:
\begin{itemize}
    \item \textbf{CSD-C/S (Character Style Distance) $\uparrow$:} Measures artistic style consistency using a style-tuned encoder.
    \item \textbf{OCCM (Occurrence Match) $\uparrow$:} Detects if the required characters appear in the shot (Recall based on Object Detection).
    \item \textbf{CP (Copy-Paste Score) $\downarrow$:} Penalizes lazy generation where the model pixel-wise copies the reference image.
    \item \textbf{Inc (Inception Score) $\uparrow$:} Measures the diversity and quality of generated assets.
    \item \textbf{Scene $\uparrow$:} Does the environment match the text description?
    \item \textbf{Character Action) $\uparrow$:} Is the character performing the specific action described?
    \item \textbf{Camera $\uparrow$:} Does the shot type (e.g., Close-up, Wide) match the instruction?
\end{itemize}

\noindent\textbf{E. Narrative Effectiveness Score (NES) - \textit{Novel Metric}}
To evaluate the Narration-Visual Synergy—a core contribution of MUSE—we introduce the Narrative Effectiveness Score (NES). This metric penalizes ``tautological narration'' (describing exactly what is seen) and rewards ``complementary narration'' (adding information not visible).

The NES is a weighted aggregation of three sub-metrics:
\begin{equation}
    NES = w_1 \cdot (G \times 5) + w_2 \cdot S + w_3 \cdot A
\end{equation}
Where weights are set to $w_1=0.3$ (Grounding), $w_2=0.4$ (Synergy), and $w_3=0.3$ (Atmosphere).

\paragraph{1. Visual Grounding ($G \in [0, 1]$).}
Measures truthfulness. The VLM starts at 1.0 and deducts points for hallucinations.
\begin{itemize}
    \item 1.0: Flawless match.
    \item 0.0: Direct contradiction (e.g., Audio says ``Day'', Visual shows ``Night'').
\end{itemize}

\paragraph{2. Information Synergy ($S \in [1, 5]$).}
Measures the ``Value Add'' of the audio channel using a Film Theory rubric.
\begin{itemize}
    \item 1 (Tautology): Narration repeats the visual (e.g., ``A man walks''). \textit{Failure.}
    \item 3 (Anchorage): Narration identifies names or specific details not fully visible. \textit{Baseline.}
    \item 5 (Counterpoint/Subtext): Audio reveals deep thematic truth or contrast that recontextualizes the image (e.g., Visual: Peaceful party; Audio: ``It was their last smile before the war'').
\end{itemize}

\paragraph{3. Atmosphere Match ($A \in [1, 5]$).}
Evaluates the pacing and tonal alignment at the story level.
\begin{itemize}
    \item 1 (Broken): Tone implies a different genre than visuals.
    \item 5 (Immersive): Perfect stylistic match; narration pauses for visual impact and speeds up for action.
\end{itemize}

\noindent{Judge Prompt for NES (Abbreviated).}
\begin{lstlisting}[language=json]
{
  "task": "Evaluate Audio-Visual Relationship",
  "rubric_synergy": {
    "1_point": "Tautology. Narration just describes the image.",
    "3_points": "Anchorage. Identifies characters/details.",
    "5_points": "Counterpoint. Adds invisible tension or thematic depth."
  },
  "rubric_grounding": {
    "instruction": "Start at 1.0. Deduct 0.3 for vague mismatch, 1.0 for hallucination."
  },
  "constraint": "If Grounding < 0.5 (Hallucination), Synergy is capped at 2."
}
\end{lstlisting}

\noindent\textbf{F. Audio Evaluation Framework}
\label{app:audio_metrics}

To rigorously evaluate the Vocal Trait Synthesis (VTS) module, we employ a streamlined ``LLM-as-a-Judge'' pipeline using \texttt{Gemini-2.5-Pro}. We focus on four critical dimensions that define character identity and production quality: \textbf{Age}, \textbf{Emotion}, \textbf{Prosody}, and \textbf{Clarity}.

\noindent{Evaluation Dimensions}
Age Consistency (Identity): Evaluates whether the synthesized voice accurately reflects the target biological age defined in the character profile (e.g., ``42 years old'' vs. ``Child'').
Emotional Integrity (Performance): Measures the alignment between the vocal affect (Valence/Arousal) and the narrative context (e.g., ``weary'', ``commanding'', ``joyful'').
Prosodic Alignment (Style): Assesses the rhythm, pacing, and cadence of the speech (e.g., ``measured and calm'' vs. ``rapid and anxious'').
Clarity (Quality): A reference-free metric evaluating speech intelligibility, signal-to-noise ratio, and the absence of robotic artifacts or distortion. We provide the exact system prompts used to instruct the VLM judge. The evaluation is conducted in two stages: Semantic Alignment (for attributes) and Acoustic Quality (for clarity).

\noindent{Stage 1: Semantic Alignment Judge.}
This agent compares the synthesized audio against the character's textual description.

\begin{lstlisting}[language=json]
{
  "task": "Evaluate Audio against Character Profile",
  "input": {
    "audio": "<audio_syn_base64>",
    "target_profile": {
      "Age": "42 years old",
      "Emotion": "Gravelly, authoritative but weary",
      "Prosody": "Measured, calm, deliberate cadence"
    }
  },
  "evaluation_criteria": {
    "Age_Match": "Does the voice sound like a 40s male?",
    "Emotion_Match": "Is the 'weary authority' audible?",
    "Prosody_Match": "Is the pacing steady/deliberate?"
  },
  "output_format": {
    "scores": {
      "age": 4.5,      // 1-5 Scale
      "emotion": 4.0,  // 1-5 Scale
      "prosody": 5.0   // 1-5 Scale
    },
    "reasoning": "Perfectly captures the measured cadence. Age sounds correct."
  }
}
\end{lstlisting}

\noindent{Stage 2: Acoustic Fidelity Judge.}
This agent blindly assesses the technical quality of the generation.

\begin{lstlisting}[language=json]
{
  "task": "Assess Audio Fidelity",
  "input_modality": ["audio_syn"],
  "focus_metric": "Clarity",
  "scoring_rubric": {
    "5 (Professional)": "Crystal clear, studio quality, perfect intelligibility.",
    "3 (Acceptable)": "Understandable but minor noise or robotic texture.",
    "1 (Unusable)": "Muffled, distorted, or incoherent speech."
  },
  "output": {
    "clarity_score": 4.8,
    "issues_detected": []
  }
}
\end{lstlisting}

\section{Additional Analysis}
\noindent\textbf{Visualization of MUSEBench.} MUSEBench encompasses prompts for five genres of stories, with the number of protagonists per story ranging from 1 to 3 (i.e., difficulty escalating from easy to hard). Figures \ref{fig:supp_musebench} present selected results generated by MUSE, which are solely used for visualizing the story scope and the number of characters.
\begin{figure}[h]
    \centering
    \includegraphics[width=0.99\linewidth]{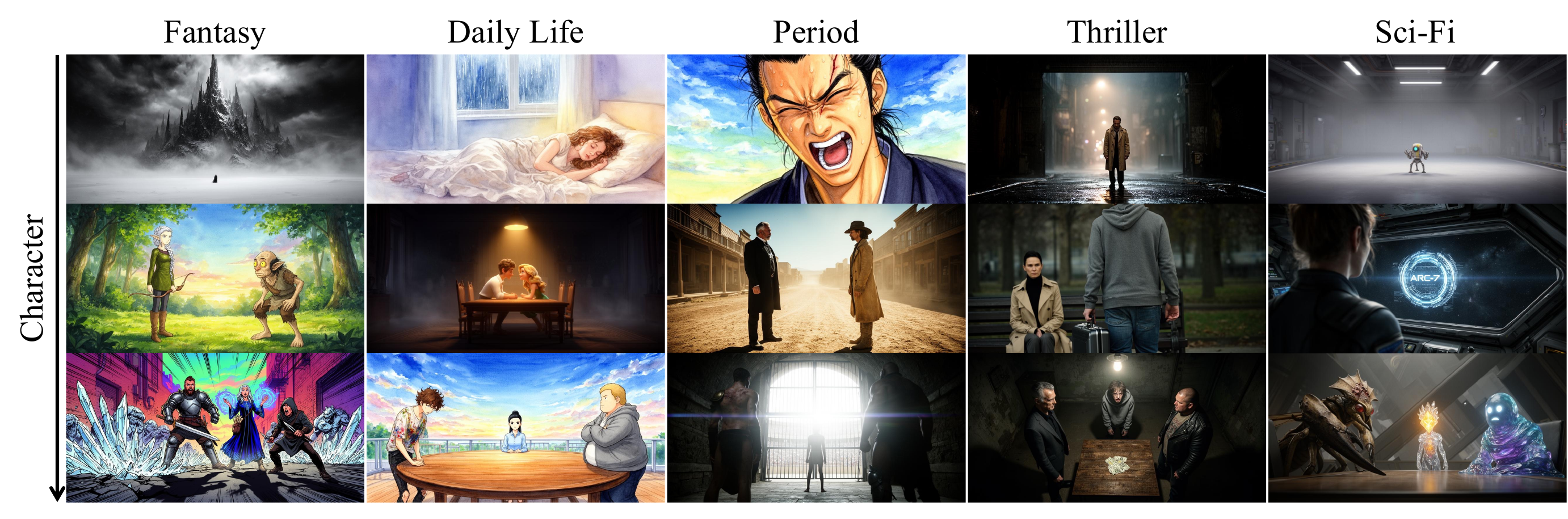}
    \vspace{-2mm}
    \caption{Visualization of MUSEBench.}
    \label{fig:supp_musebench}
    \vspace{-2mm}
\end{figure}

\noindent\textbf{Character Assets.} The results generated by MUSE depend on high-quality character assets. To adapt to diverse camera movements, MUSE requires images of the same character from multiple viewpoints. Figure \ref{fig:supp_characterassets} present the character assets produced by MUSE. To eliminate the "texture-mapping look" of scene images, MUSE leverages RMBG for background removal. Additionally, it utilizes VLMs to generate assets from various viewpoints.
\begin{figure}[h]
    \centering
    \includegraphics[width=0.99\linewidth]{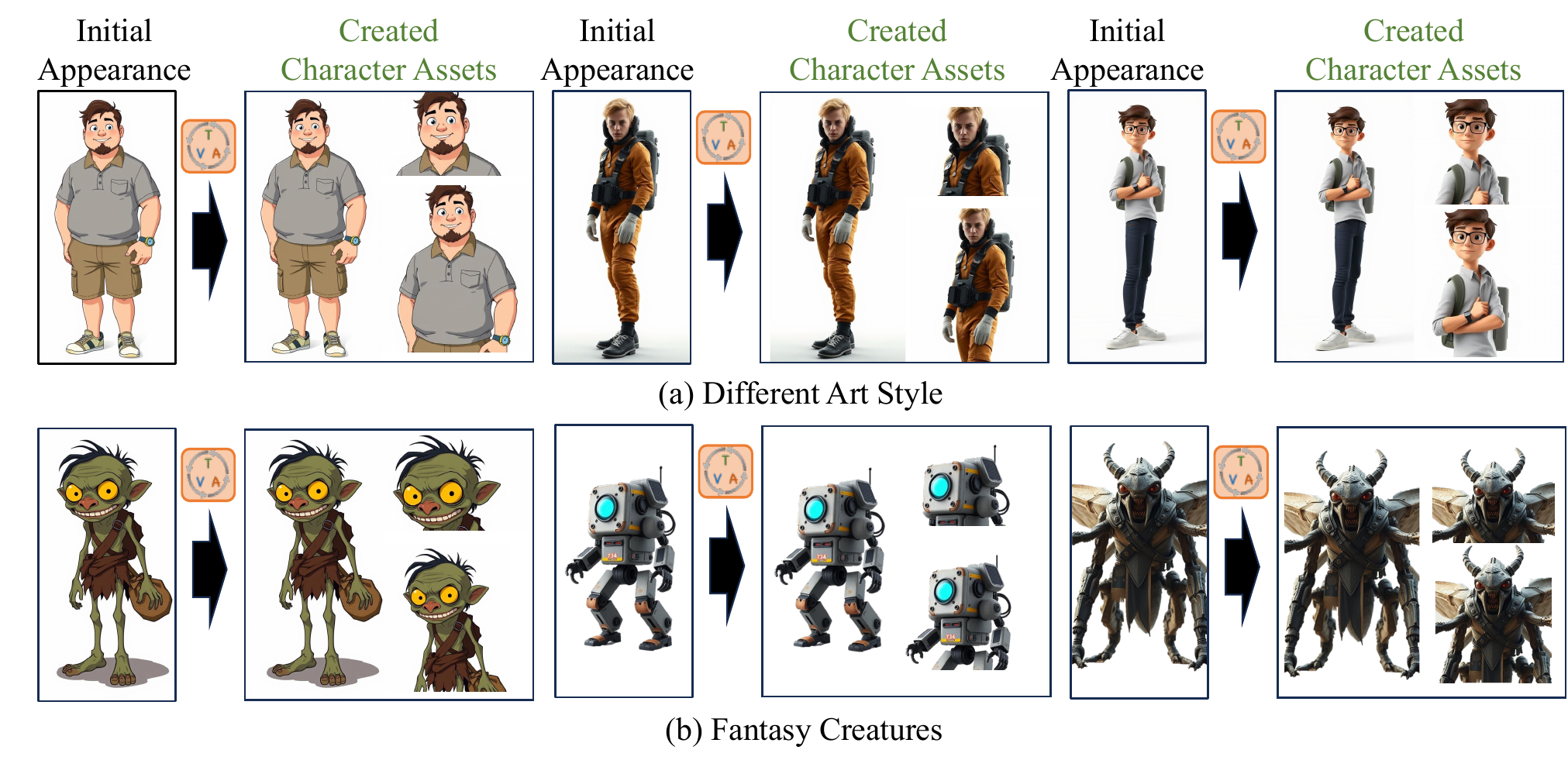}
    \vspace{-2mm}
    \caption{Visualization of Generated Character Assets.}
    \label{fig:supp_characterassets}
    \vspace{-2mm}
\end{figure}

\noindent\textbf{Style Anchor from LLM.} A persistent challenge in long-form video generation is Visual Style Drift. MUSE addresses this through a rigorous Global Style Anchoring mechanism. MUSE leverages LLMs to analyze the script and select the most suitable style from the built-in style library, which is employed as a Style Anchor throughout the entire generation process. (It is worth noting that this parameter can be manually configured; in this work, we opt for the agent to autonomously make the selection. While this increases the generation complexity, we believe that results with diverse styles are more engaging.) We define a strict schema for visual styles. Below is the data structure defining a style preset (e.g., Pixar-3D), where character and scene prompts are decoupled to ensure flexibility.

\noindent{Visual Style Data Structure}
\begin{lstlisting}[language=json]
class VisualStyle:
    name: str
    display_name: str  
    description: str   
    char_prompt: str   
    scene_prompt: str  
    negative_prompt: str

# Example: Pixar 3D Preset
PIXAR_STYLE = VisualStyle(
    name="pixar_3d",
    display_name="3D Animation (Pixar Style)",
    description=(
        "Family friendly, rounded features, soft lighting, "
        "vibrant colors, expressive character design."
    ),
    char_prompt=(
        "3D animated style, Pixar quality, soft studio lighting, "
        "c4d render, unreal engine 5"
    ),
    scene_prompt=(
        "3D rendered environment, stylized textures, "
        "volumetric fog, soft ambient lighting"
    ),
    negative_prompt="2d, sketch, realistic, photograph, lowres"
)
\end{lstlisting}

\noindent{Intelligent Style Selection}
Before generation, MUSE selects the optimal Style Anchor based on narrative tone.

\noindent{Style Analysis Output}
\begin{lstlisting}[language=json]
{
  "script_analysis": {
    "genre": "Slice of Life",
    "tone": "Melancholic but hopeful",
    "audience": "Adult"
  },
  "decision": {
    "selected_style_id": "watercolor",
    "reasoning": [
      "The script emphasizes introspection and mood over action.",
      "The 'Watercolor' style's soft edges and bleed effects",
      "mirror the rainy atmosphere better than photorealism."
    ]
  }
}
\end{lstlisting}

\noindent{Universal Style Injection}
Once selected, the style becomes an Immutable Constraint. The system enforces consistency by injecting style prompts at the very beginning of every generation request.

\noindent{Prompt Injection Logic}
\begin{lstlisting}[language=json]
def to_image_prompt(self, style_modifier: str) -> str:
    prompt_parts = [
        # 1. Hard Constraint: Style Anchor (Priority High)
        style_modifier, 
        
        # 2. Hard Constraint: Framing (Full Body)
        "FULL BODY PORTRAIT, Single Character",
        
        # 3. Variable Content: Character Details
        f"a {self.age} character named {self.name}",
        f"{self.physical_appearance}",
        
        # 4. Quality Boosters
        "detailed, high quality"
    ]
    # Join parts with commas to form the final Flux prompt
    return ", ".join(prompt_parts)
\end{lstlisting}

\noindent Figure \ref{fig:supp_style} shows the results of different styles generated by MUSE. Thanks to the Style Anchor, MUSE can generate results with diverse yet consistent styles.

\begin{figure}[h]
    \centering
    \includegraphics[width=0.99\linewidth]{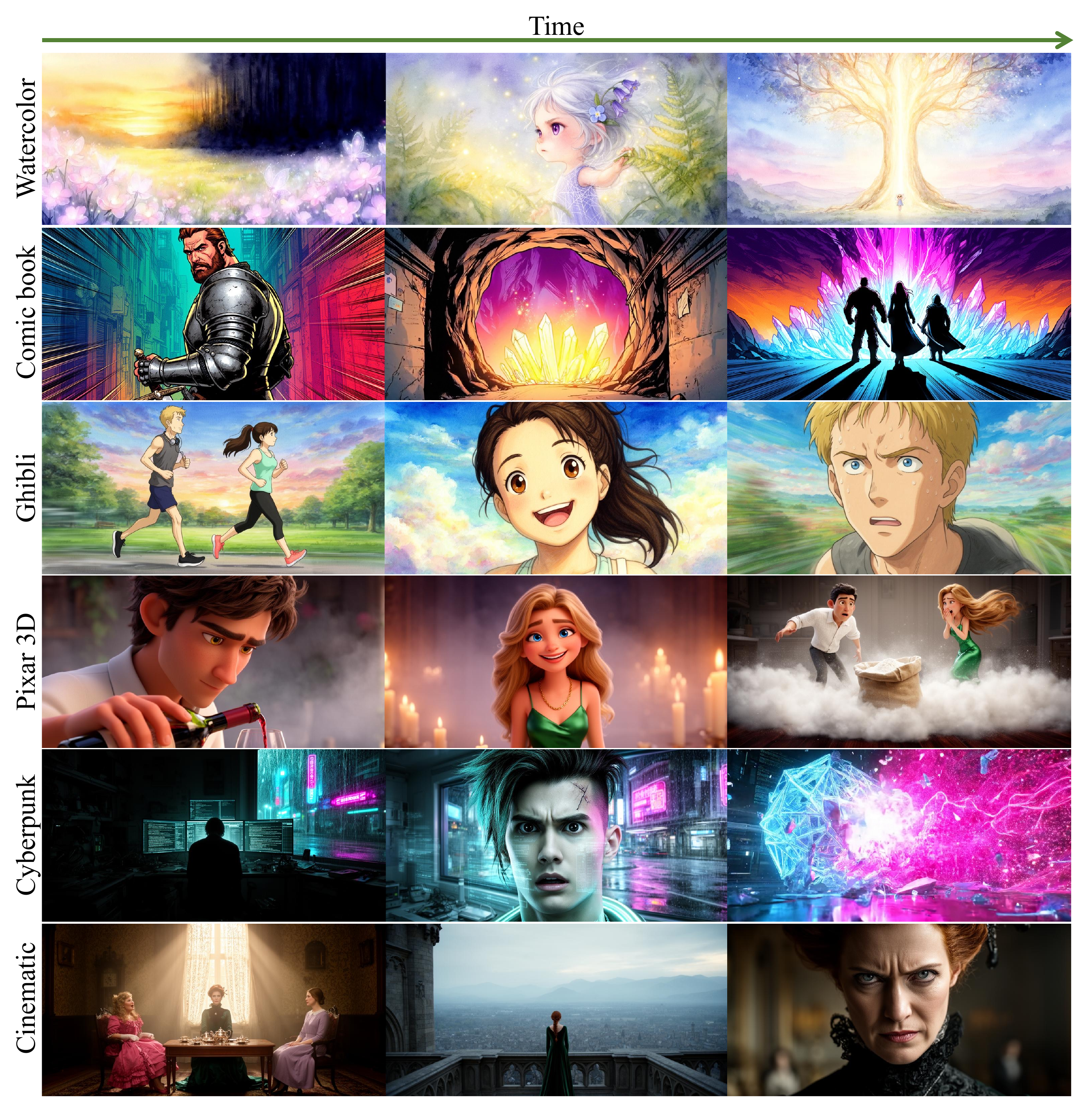}
    \vspace{-2mm}
    \caption{Visualization of Results in Different Styles.}
    \label{fig:supp_style}
    \vspace{-2mm}
\end{figure}

\noindent\textbf{Failure Cases.} Beyond the cases discussed in the main text, identity inconsistency tends to occur during shot transitions when the character is a fantasy creature (Figure \ref{fig:supp_failure}). Future work will include more refined generation strategies tailored to fantasy creatures.
\begin{figure}[h]
    \centering
    \includegraphics[width=0.99\linewidth]{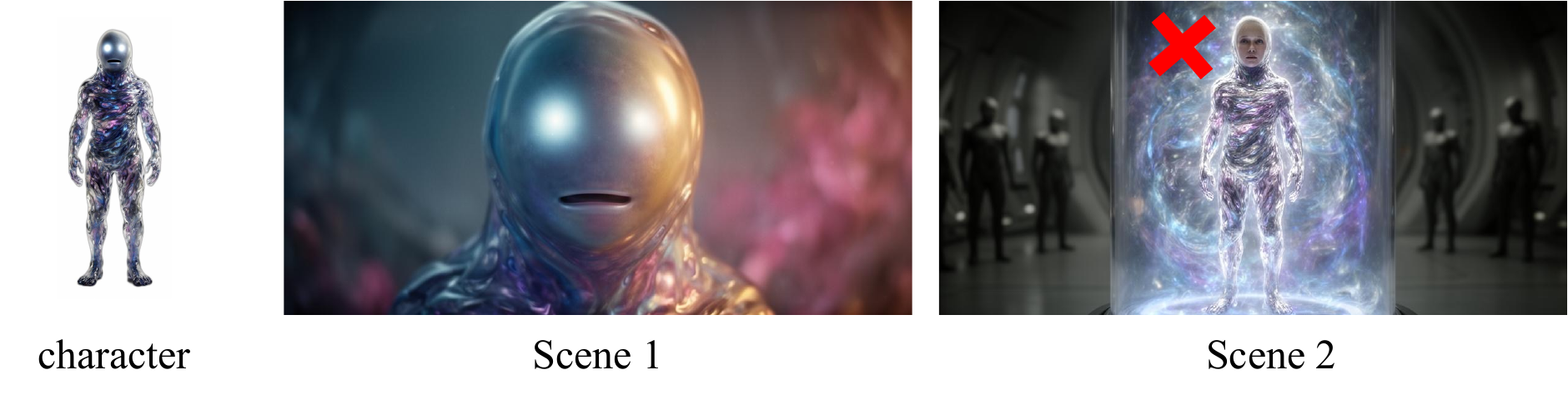}
    \vspace{-2mm}
    \caption{Failure case on fantasy creature.}
    \label{fig:supp_failure}
    \vspace{-2mm}
\end{figure}

%% file: tables/agentoverview.tex
\begin{table}[h]
\centering
\caption{MUSE Agent Specifications. A compact view of the controller $\mathcal{M}$ invoking specialist agents. Input states are transformed into structured outputs or error vectors.}
\label{tab:agent_overview}
\resizebox{1.0\linewidth}{!}{
\begin{tabular}{l|l|l}
\toprule
\textbf{Agent} & \textbf{I/O Mapping} & \textbf{Functional Purpose} \\
\midrule
\multicolumn{3}{l}{\textit{\textbf{Phase 1: Pre-production (Identity Anchoring)}}} \\
Screenwriter & $\mathcal{U} \to \mathcal{S}$ & Expand user prompt into structured script. \\
Planner & $\mathcal{S} \to \{\text{cast, shots}\}$ & Decompose script into atomic shot specs. \\
VTS Module & $\text{traits} \to \mathbf{z}_{voc}$ & Synthesize zero-shot acoustic anchor. \\
Visual Casting & $\text{traits} \to \mathbf{z}_{vis}$ & Generate consistent visual reference sheet. \\
Critic ($\Psi_{pre}$) & $(\text{ref}, \mathcal{S}) \to \mathbf{e}$ & Verify cross-modal age/gender alignment. \\
\midrule
\multicolumn{3}{l}{\textit{\textbf{Phase 2: Production (Spatial Composition)}}} \\
Router & $(s_i, \mathcal{H}) \to m$ & Select optimal path (Text-to-Video vs. I2V). \\
LayoutGen & $s_i \to L^{(i)}_{bbox}$ & Generate coarse spatial bounding boxes. \\
Asset Gen & $(\Theta, L, \mathbf{z}_{vis}) \to x$ & Synthesize assets under identity constraints. \\
Critic ($\Psi_{prod}$) & $(x, L, s_i) \to \mathbf{e}$ & Detect artifacts and layout violations. \\
\midrule
\multicolumn{3}{l}{\textit{\textbf{Phase 3: Post-production (Temporal Synthesis)}}} \\
Action Planner & $(s_i, \mathcal{H}) \to \text{cm}$ & Define camera motion and boundaries. \\
VideoGen & $(\Theta, \text{Tail}(v_{i-1})) \to v_i$ & Autoregressive chunk generation. \\
Critic ($\Psi_{post}$) & $(v_i, s_i) \to \mathbf{e}$ & Audit boundary compliance and fluidity. \\
\bottomrule
\end{tabular}
}
\end{table}